\documentclass[sigconf]{acmart}
\usepackage{graphicx}
\usepackage{booktabs} % for professional tables
\usepackage{multirow}
\usepackage{threeparttable}
\usepackage{verbatim}
\usepackage{multicol}
\usepackage{algorithm}
\usepackage{algorithmic}
\usepackage{underscore}
\usepackage{bm}

\copyrightyear{2021}
\acmYear{2021}
\setcopyright{acmcopyright}\acmConference[CIKM '21]{Proceedings of the 30th ACM International Conference on Information and Knowledge Management}{November 1--5, 2021}{Virtual Event, QLD, Australia}
\acmBooktitle{Proceedings of the 30th ACM International Conference on Information and Knowledge Management (CIKM '21), November 1--5, 2021, Virtual Event, QLD, Australia}
\acmPrice{15.00}
\acmDOI{10.1145/3459637.3482248}
\acmISBN{978-1-4503-8446-9/21/11}

\begin{document}
\fancyhead{}
\title{Delve into the Performance Degradation of \\
Differentiable Architecture Search}


\author{Jiuling Zhang}
\email{zhangjiuling19@mails.ucas.ac.cn}
\affiliation{%
  \institution{University of Chinese Academy of Sciences}
  \city{Beijing}
  \country{China}}

\author{Zhiming Ding}
\authornote{Corresponding author.}
\email{zhiming@iscas.ac.cn}
\affiliation{%
  \institution{Institute of Software Chinese Academy of Sciences}
  \city{Beijing}
  \country{China}}


\begin{abstract}
Differentiable architecture search (DARTS) is widely considered to be easy to overfit the validation set which leads to performance degradation. We first employ a series of exploratory experiments to verify that neither high-strength architecture parameters regularization nor warmup training scheme can effectively solve this problem. Based on the insights from the experiments, we conjecture that the performance of DARTS does not depend on the well-trained supernet weights and argue that the architecture parameters should be trained by the gradients which are obtained in the early stage rather than the final stage of training. This argument is then verified by exchanging the learning rate schemes of weights and parameters. Experimental results show that the simple swap of the learning rates can effectively solve the degradation and achieve competitive performance. Further empirical evidence suggests that the degradation is not a simple problem of the validation set overfitting but exhibit some links between the degradation and the operation selection bias within bilevel optimization dynamics. We demonstrate the generalization of this bias and propose to utilize this bias to achieve an operation-magnitude-based selective stop.
\end{abstract}

\begin{CCSXML}
<ccs2012>
<concept>
<concept_id>10010147.10010178.10010205</concept_id>
<concept_desc>Computing methodologies~Search methodologies</concept_desc>
<concept_significance>500</concept_significance>
</concept>
<concept>
<concept_id>10010147.10010257.10010293.10010294</concept_id>
<concept_desc>Computing methodologies~Neural networks</concept_desc>
<concept_significance>500</concept_significance>
</concept>
</ccs2012>
\end{CCSXML}

\ccsdesc[500]{Computing methodologies~Search methodologies}
\ccsdesc[500]{Computing methodologies~Neural networks}

\keywords{auto deep learning; neural architecture search; }


\maketitle

\section{Introduction}
Evaluating by training from scratch is the key bottleneck of the neural architecture search (NAS). Gradient-based methods only need to train a single large-scale network (supernet) once and for all, which are demonstrably faster than the discrete optimization counterparts \cite{zoph2016neural,zoph2018learning}, thereby becoming a highly promising branch of the NAS research under limited budget.
Gradient-based methods are basically divided into two research directions: 1. One-shot NAS; 2. Differentiable NAS. Inspired by the successful practices of the weight sharing paradigm in transfer learning and multi-task learning, One-shot NAS \cite{bender2018understanding,dong2019one} proposes to train a supernet subsuming all possible architectures with compound edges in the search space. Individual performance can then be evaluated by inheriting the corresponding weights from the supernet and performing a single feedforward.

Differentiable architecture search comes from the proposal of DARTS \cite{liu2018darts} which is inspired by the differentiable relaxation of the discrete selection based on the attention mechanism. DARTS places a mixture of operation candidates $O = \left\{ {o_{i,j}^1,o_{i,j}^2,...,o_{i,j}^M} \right\}$ on each edge and continuously relaxes the categorical operation selection to a softmax over all operation candidates as shown in Eq.(1).
\begin{equation}
categorical(O) \approx softmax (A) = \frac{{\exp (\alpha _{i,j}^o)}}{{\sum\nolimits_{o' \in O} {\exp (\alpha _{i,j}^{o'})} }}
\end{equation}
\begin{equation}
\sum\limits_{m = 1}^M {categorical(O)}\ o_{i,j}^m({x_i}) \approx \sum\limits_{m = 1}^M {softmax (A)}\  o_{i,j}^m({x_i})
\end{equation}
where $A = \left\{ {\alpha _{i,j}^1,\alpha _{i,j}^2,...,\alpha _{i,j}^M} \right\}$ is a set of architecture parameters corresponding to the operation set $O$. $x_i$ is the input feature maps of the operation $o_{i,j}$. Henceforth, we abbreviate supernet weights as weights and architecture parameters as parameters.

\begin{figure}[ht]
\begin{center}
\includegraphics[width=\columnwidth]{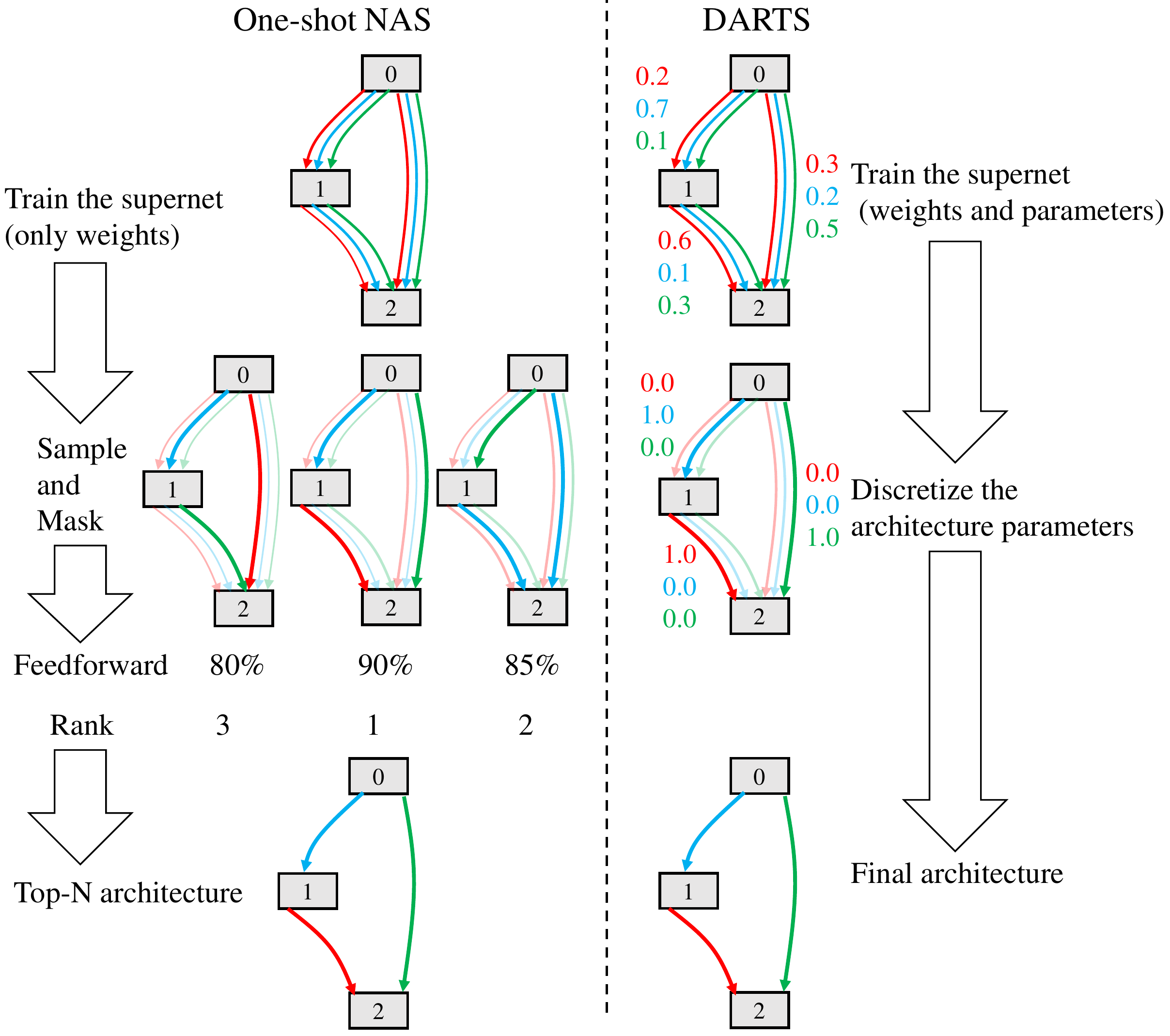}
\caption{A conceptual visualization for the differences between One-shot NAS and DARTS.}
\label{fig1}
\end{center}
\end{figure}

The most significant difference between DARTS and One-shot NAS is that DARTS explicitly parameterizes the neural architecture through the continuous differentiable parameters and alternately optimizes parameters $\alpha$ and weights $\omega$ on training and validation set in a bilevel gradient descent training recipe depicted in Eq.(3).
\begin{equation}
\mathop {\min }\limits_\alpha  {\mathcal{L}_{val}}\left( {{\omega ^ * }\left( \alpha  \right),\alpha } \right)
\ \ s.t.\ {\omega ^ * }\left( \alpha  \right) = \arg \mathop {\min }\limits_\omega  {\mathcal{L}_{train}}\left( {\omega ,\alpha } \right)
\end{equation}
Figure~\ref{fig1} outlines the processes of One-shot NAS and DARTS. DARTS is generally faster because it directly discretizes the parameters through a predefined approach to obtain the final architecture by which DARTS eliminates both the sampling and ranking processes as shown in Figure~\ref{fig1}. Furthermore, the training scheme of DARTS is much simpler because it applies SGD to optimize the parameters by obtaining gradients from a generic loss which is used as a supervision signal for the architecture search. We refer to \cite{liu2018darts} for more details of DARTS. 

In summary, One-shot NAS decidedly relies on the well-trained uncoadapted weights, otherwise the performance of ranking will be remarkably degraded \cite{bender2018understanding,pourchot2020share,zhang2020deeper}. In contrast, the performance of DARTS depends on the validity of the gradients which indicates the following two things:
\begin{itemize}
\setlength{\itemsep}{0pt}
\setlength{\parsep}{0pt}
\setlength{\parskip}{0pt}
\item Well-trained weights are not inherently necessary for DARTS to obtain the informative gradients of the parameters;
\item The performance of DARTS is affiliated with the SGD optimization dynamics of the neural network.
\end{itemize}
The research on the optimization dynamics is still shallow for the neural network with practical depth and even further, according to Eq.(3), the optimization of weights $\omega$ and parameters $\alpha$ are highly entangled due to the alternate training recipe. In this paper, by delving into the performance degradation of DARTS, we finally focus on the interaction between weights and parameters within bilevel optimization dynamics. Specifically, our main contributions can be summarized as follows:
\begin{itemize}
\setlength{\itemsep}{0pt}
\setlength{\parsep}{0pt}
\setlength{\parskip}{0pt}
\item We empirically study the performance degradation and theoretically analyse the impact of the parameter learning rate scheme in DARTS. Afterwards we provide the easiest way to solve the performance degradation by exchanging the learning rate schemes between weights and parameters;
\item We design some experiments to manifest that the over-trained weights play an important role in degradation. We also demonstrate that the degradation is not a simple problem of the validation set overfitting but link to some specific operations in the search space;
\item By introducing the concept of the operation magnitude, we uncover the operation selection bias via studying the
parameter optimization dynamics. We further propose the operation-magnitude-based selective stop to eliminate the performance degradation of DARTS;
\item By conducting extensive experiments on multiple datasets and search spaces, we demonstrate the efficacy of our methods comparing to some well-developed stop criteria and strong baselines.
\end{itemize}

\section{Methodology}
Due to the budget of training and evaluation of DARTS \cite{yang2019evaluation}, we choose NAS-BENCH-201 \cite{dong2019bench} as the main experimental and verification platform. NAS-BENCH-201 can stably reproduce the performance degradation of DARTS under the default settings and more importantly, it makes the study of bilevel optimization dynamic computationally affordable. In particular, unlike many other DARTS studies, NAS-BENCH-201 employs different seeds by default which is more conducive to obtain some general patterns independent of the specific initialization. We follow this rule and take the average after five repeats under different seeds. We refer to \cite{dong2019bench} for more information of NAS-BENCH-201. We detail all the experimental setting changes within each experimental description.
\subsection{Exploratory Experiments}
\begin{figure*}[ht]
\begin{center}
\centerline{\includegraphics[width=\linewidth]{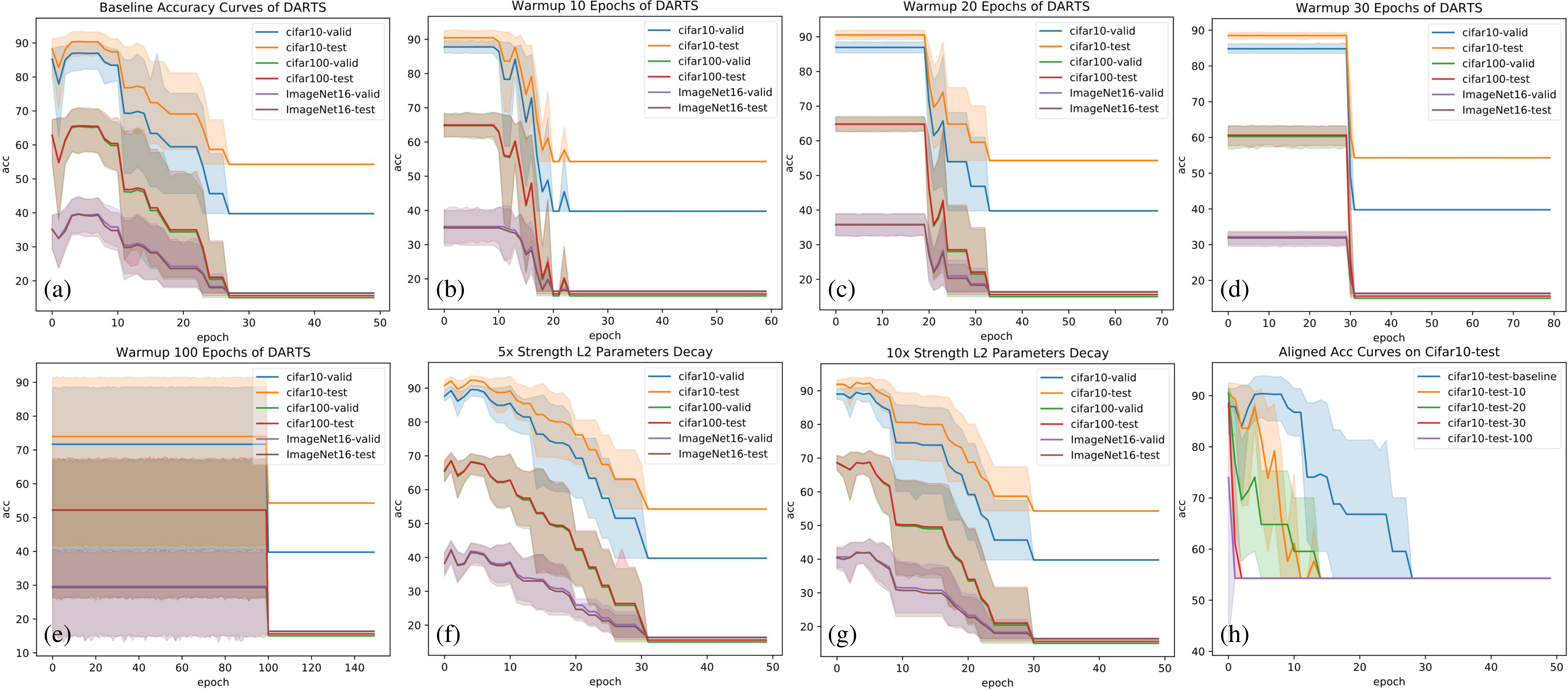}}
\caption{(a) Baseline; (b)\textasciitilde(d) Results of Exp.1; (f)\textasciitilde(g) Results of Exp.2; (e) Results of Exp.3; (h) The aligned accuracy curves of the last 50 epochs in baseline, Exp.1 and Exp.3 respectively. }
\label{fig2}
\end{center}
\begin{center}
\centerline{\includegraphics[width=\linewidth]{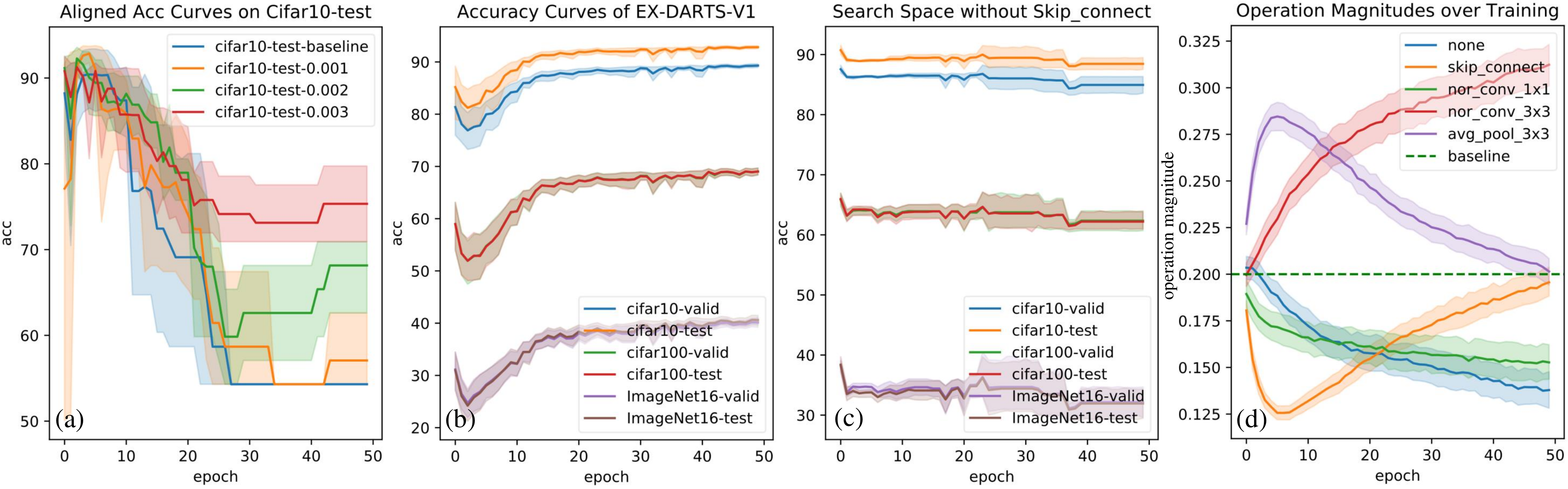}}
\caption{Accuracy curves of (a) Exp.4; (b) Exp.5; (c) Exp.6. (d) Operation magnitudes of Exp.5.}
\label{fig3}
\end{center}
\end{figure*}

We first set the searching curves of DARTS-V1 under the default settings on NAS-BENCH-201 as the baseline shown in Figure~\ref{fig2}(a). Previous studies claim that DARTS can restrain the randomness of parameter updates and it is prone to choose low-computational operations due to insufficient training of weights \cite{wu2019fbnet,yan2019hm}. We can accordingly derive an hypothesis (\textbf{HP.1}) that the training of parameters can benefit from the well-trained weights. Based on HP.1, many researches employ two common solutions on DARTS :

\textit{\textbf{S.1}: Special learning rate schemes with exceptional small parameter learning rate (0.0003) and normal weight learning rate (0.025);}

\textit{\textbf{S.2}: Warmup training scheme.}

Since S.1 has been employed by default on NAS-BENCH-201, we focus on verifying whether the warmup training scheme helps to alleviate the performance degradation. Warmup is a widely used supernet training scheme \cite{xu2019pc,yan2019hm,vahdat2020unas,chen2019progressive,wu2019fbnet} that the parameters are frozen at the beginning for several epochs while updating weights independently to warmup the weights. Experiment N is abbreviated as Exp.N in the rest of the paper. The first experiment can be summarized as follows:

\textit{\textbf{Exp.1}: Test 10, 20 and 30 warmup epochs respectively. For a fair comparison, both weights and parameters are then jointly trained enough for another 50 epochs according to the default settings.}

In addition, we also verify whether the general regularization for the parameters is helpful since previous studies also claim that the degradation is due to the parameter overfitting of the validation set \cite{arber2020understanding,xu2019pc}. L2 decay is the most common regularization approach which is already included in the default settings, our experiments are to test the effectiveness of the high-strength L2 regularization for the parameters. We follow the advice mentioned in NAS-BENCH-201 to avoid overfitting by not regularizing the parameters of a specific operation but ensuring that all parameters are regularized at the same strength. Exp.2 can be summarized as follows: 

\textit{\textbf{Exp.2}: Test the L2 parameters regularization with intensity of 0.005 and 0.01 respectively, which are 5$\times$ and 10$\times$ higher than the default value (0.001).}

We further test well-trained weights beyond the normal warmup scheme as a sanity check in Exp.3.

\textit{\textbf{Exp.3}: Train the weights for 100 epochs while freezing the parameters. Then both weights and parameters are updated for another 50 epochs under the default settings.}

Figure~\ref{fig2}(b)\textasciitilde(g) exhibit the accuracy curves of DARTS over the training epochs in the exploratory experiments. Figure~\ref{fig2}(h) shows the aligned curves of the final 50 epochs in baseline, Exp.1 and Exp.3 respectively. We have the following observation summaries (\textbf{OS}) based on the exploratory experimental results:

\textit{\textbf{OS.1}: Figure~\ref{fig2}(b)\textasciitilde(g) demonstrate that neither warmup training scheme nor high-strength parameter regularization can effectively address the performance degradation of DARTS;}

\textit{\textbf{OS.2}: We can see from 
Figure~\ref{fig2}(h) that when the warmup epochs are extended, the performance of DARTS degrades faster after the parameters are unfrozen which is contrary to HP.1 and somewhat counterintuitive.}

\begin{comment}
(more in Appendix C)
\end{comment}

Particularly shown in Figure~\ref{fig2}(e), DARTS almost immediately suffers from the degradation in Exp.3 as soon as the parameters are unfrozen after weights are pretrained alone for 100 epochs.

\subsection{Exchange Learning Rate Schemes}
Based on the insights from the exploratory experiments, we find that the well-trained weights are not beneficial to the training of parameters, thus the warmup scheme (S.2) proposed under HP.1 cannot solve the performance degradation. Accordingly, we have reason to suspect that the learning rate schemes (S.1) based on the same HP.1 may not work as well. In this chapter, we first break S.1 by moderately increasing the parameter learning rate and observe its impact on DARTS which can be summarized as follows:

\textit{\textbf{Exp.4}: Test the performance of DARTS when the parameter learning rate is 0.001, 0.002, and 0.003 respectively, corresponding to 3.3$\times$, 6.6$\times$ and 10$\times$ of the exceptional small parameter learning rate (0.0003) under the default settings.}

Figure~\ref{fig3}(a) illustrates the results of Exp.4 from which we can find OS.3 as follows:

\textit{\textbf{OS.3}: The performance degradation of DARTS is gradually alleviated by increasing the parameter learning rate.}

To explain OS.3, we next theoretically investigate the impact of increasing parameter learning rate on DARTS through both qualitative analysis and numerical study. Softmax normalizes the input vector ${\bm{x}} = \{ {x_1},...,{x_d}\}$ to a probability distribution ${\bm{y}} = \{ {y_1},...,{y_d}\}$ by ${y_i} = {\raise0.7ex\hbox{${\exp ({x_i})}$} \!\mathord{\left/
 {\vphantom {{\exp ({x_i})} {\sum\nolimits_{j = 1}^d {{x_j}} }}}\right.\kern-\nulldelimiterspace}
\!\lower0.7ex\hbox{${\sum\nolimits_{j = 1}^d {{\exp (x_j)}} }$}}$.
Obviously, softmax is a multivariate function that the derivative needs to be discussed separately for each output entry ${y_j}$ with respect to input entry ${x_i}$ as shown in Eq.(4).
\begin{equation}
\frac{{\partial {y_j}}}{{\partial {x_i}}} = \left\{ {\begin{array}{*{20}{c}}
\begin{aligned}
&{y_i}\left( {1 - {y_i}} \right)&{\rm{for}}\ i = j\\
& - {y_j}{y_i}&{\rm{for}}\ i \ne j
\end{aligned}
\end{array}} \right.
\end{equation}
By combining the two cases in Eq.(4), we can get the Jacobian matrix of softmax in Eq.(5).
\begin{equation}
\frac{{\partial {\bm{y}}}}{{\partial {\bm{x}}}} = \left[ {\begin{array}{*{20}{c}}
{{y_1} - y_1^2}&{ - {y_1}{y_2}}&{ - {y_1}{y_3}}& \cdots &{ - {y_1}{y_d}}\\
{ - {y_2}{y_1}}&{{y_2} - y_2^2}&{ - {y_2}{y_3}}& \cdots &{ - {y_2}{y_d}}\\
 \vdots & \vdots & \vdots & \cdots & \vdots \\
{ - {y_d}{y_1}}&{ - {y_d}{y_2}}&{ - {y_d}{y_3}}& \cdots &{{y_d} - y_d^2}
\end{array}} \right]
\end{equation}
Let $l$ be the loss function then we can get ${\raise0.7ex\hbox{${\partial l}$} \!\mathord{\left/
 {\vphantom {{\partial l} {\partial {\bm{x}}}}}\right.\kern-\nulldelimiterspace}
\!\lower0.7ex\hbox{${\partial {\bm{x}}}$}}$ as Eq.(6).

\begin{equation}
\resizebox{\columnwidth}{!}{$
\frac{{\partial {\bm{y}}}}{{\partial {\bm{x}}}}\frac{{\partial l}}{{\partial {\bm{y}}}} = \left[ {\begin{array}{*{20}{c}}
{{y_1} - y_1^2}&{ - {y_1}{y_2}}&{ - {y_1}{y_3}}& \cdots &{ - {y_1}{y_d}}\\
{ - {y_2}{y_1}}&{{y_2} - y_2^2}&{ - {y_2}{y_3}}& \cdots &{ - {y_2}{y_d}}\\
 \vdots & \vdots & \vdots & \cdots & \vdots \\
{ - {y_d}{y_1}}&{ - {y_d}{y_2}}&{ - {y_d}{y_3}}& \cdots &{{y_d} - y_d^2}
\end{array}} \right]\left[ {\begin{array}{*{20}{c}}
{{{\partial l} \mathord{\left/
 {\vphantom {{\partial l} {\partial {y_1}}}} \right.
 \kern-\nulldelimiterspace} {\partial {y_1}}}}\\
{{{\partial l} \mathord{\left/
 {\vphantom {{\partial l} {\partial {y_2}}}} \right.
 \kern-\nulldelimiterspace} {\partial {y_2}}}}\\
 \vdots \\
{{{\partial l} \mathord{\left/
 {\vphantom {{\partial l} {\partial {y_d}}}} \right.
 \kern-\nulldelimiterspace} {\partial {y_d}}}}
\end{array}} \right]$
}
\end{equation}
Take the absolute value of the partial derivative of the output vector ${\bm{y}}$ with respect to each input entry $x_i$ as \[\left| {{\raise0.7ex\hbox{${\partial {\bm{y}}}$} \!\mathord{\left/
 {\vphantom {{\partial {\bm{y}}} {\partial {x_i}}}}\right.\kern-\nulldelimiterspace}
\!\lower0.7ex\hbox{${\partial {x_i}}$}}} \right| = \left[ {\begin{array}{*{20}{c}}
{{y_i}{y_1}}& \cdots &{{y_i} - y_i^2}& \cdots &{{y_i}{y_d}}
\end{array}} \right]\]
We next take partial derivative again of $\left| {{\raise0.7ex\hbox{${\partial {\bm{y}}}$} \!\mathord{\left/
 {\vphantom {{\partial {\bm{y}}} {\partial {x_i}}}}\right.\kern-\nulldelimiterspace}
\!\lower0.7ex\hbox{${\partial {x_i}}$}}} \right|$ with respect to ${y_i}$. For the entry where $i = j$, we can get Eq.(7).
\begin{equation}
\frac{{\partial ({y_i} - y_i^2)}}{{\partial {y_i}}} = 1 - 2{y_i}{\rm{\ \ \ for }}\ \ i = j
\end{equation}
Let $\sum {{y_{i \ne j}}} $ denotes $\sum\limits_{j \ne i}^d {{y_j}} $, we then calculate ${\raise0.7ex\hbox{${\partial \left( {{y_i}{y_j}} \right)}$} \!\mathord{\left/
 {\vphantom {{\partial \left( {{y_i}{y_j}} \right)} {\partial {y_i}}}}\right.\kern-\nulldelimiterspace}
\!\lower0.7ex\hbox{${\partial {y_i}}$}}$ where $i \ne j$ as Eq.(8).
\begin{equation}
\begin{aligned}
\frac{{\partial \left( {{y_i}{y_j}} \right)}}{{\partial {y_i}}} &= \frac{{\partial \left( {{y_i}\left( {1 - \sum {{y_{i \ne j}}}  - {y_i} + {y_j}} \right)} \right)}}{{\partial {y_i}}}\\
&= 1 - 2{y_i} - \sum {{y_{i \ne j}}}  + {y_j}\\
&= {y_j} - {y_i}{\rm{\ \ \ for }}\ \ {i \ne j}
\end{aligned}
\end{equation}
By combining Eq.(7) and Eq.(8), we can get Eq.(9).
\begin{equation}
\frac{{\left| {{\raise0.7ex\hbox{${\partial {\bm{y}}}$} \!\mathord{\left/
 {\vphantom {{\partial {\bm{y}}} {\partial {x_i}}}}\right.\kern-\nulldelimiterspace}
\!\lower0.7ex\hbox{${\partial {x_i}}$}}} \right|}}{{\partial {y_i}}} = \left\{ {\begin{array}{*{20}{c}}
{1 - 2{y_i}{\rm{\ \ for  }}\ i = j}\\
{{y_j} - {y_i}{\rm{\ \ for  }}\ i \ne j}
\end{array}} \right.
\end{equation}
From Eq.(9) we can get Eq.(10).
\begin{equation}
\left| {\frac{{\partial {\bm{y}}}}{{\partial {x_i}}}} \right|\left\{ {\begin{array}{*{20}{c}}
\begin{aligned}
& \propto {y_i}&&{\rm{for  }}\ {y_i} < 0.5{\rm{\ and\ }}{y_i} < {y_j}\\
& \propto {\raise0.7ex\hbox{$1$} \!\mathord{\left/
 {\vphantom {1 {{y_i}}}}\right.\kern-\nulldelimiterspace}
\!\lower0.7ex\hbox{${{y_i}}$}}&&{\rm{for}}\ {y_i} > 0.5
\end{aligned}
\end{array}} \right.
\end{equation}
Suppose the training is divided into two phases lasting ${t_1}$ and ${t_2}$ epochs respectively. The values of ${{\raise0.7ex\hbox{${\partial l}$} \!\mathord{\left/
 {\vphantom {{\partial l} {\partial y}}}\right.\kern-\nulldelimiterspace}
\!\lower0.7ex\hbox{${\partial {\bm{y}}}$}}}$ remain the same in the two phases but the signs are reversed at the epoch ${t=t_1}$. Let ${{\bm{x}}^t} = \left[ {x_1^t,...,x_d^t} \right]$ indicates the input vector ${\bm{x}}$ at the epoch $t$. $x_{ \uparrow  \downarrow }^{{t_1}{t_2}}$ indicates $x \in {\bm{x}}$ that ${\raise0.7ex\hbox{${\partial l}$} \!\mathord{\left/
 {\vphantom {{\partial l} {\partial x_ \uparrow ^{{t_1}}}}}\right.\kern-\nulldelimiterspace}
\!\lower0.7ex\hbox{${\partial x_ \uparrow ^{{t_1}}}$}} > 0$ when $t < {t_1}$ and ${\raise0.7ex\hbox{${\partial l}$} \!\mathord{\left/
 {\vphantom {{\partial l} {\partial x_ \downarrow ^{{t_2}}}}}\right.\kern-\nulldelimiterspace}
\!\lower0.7ex\hbox{${\partial x_ \downarrow ^{{t_2}}}$}} < 0$ when $t \ge {t_1}$. $x_{ \downarrow  \uparrow }^{{t_1}{t_2}}$ indicates $x \in {\bm{x}}$ that ${\raise0.7ex\hbox{${\partial l}$} \!\mathord{\left/
 {\vphantom {{\partial l} {\partial x_ \downarrow ^{{t_1}}}}}\right.\kern-\nulldelimiterspace}
\!\lower0.7ex\hbox{${\partial x_ \downarrow ^{{t_1}}}$}} < 0$ when $t < {t_1}$ and ${\raise0.7ex\hbox{${\partial l}$} \!\mathord{\left/
 {\vphantom {{\partial l} {\partial x_ \uparrow ^{{t_2}}}}}\right.\kern-\nulldelimiterspace}
\!\lower0.7ex\hbox{${\partial x_ \uparrow ^{{t_2}}}$}} > 0$ when $t \ge {t_1}$. We only consider the case when the softmax input entry ${{x_i}}$ and the corresponding output entry ${{y_i}}$ are monotonic. In the training phase $t < {t_1}$, the positive gradients of $x_ \uparrow ^{{t_1}}$ lead to the rise of the corresponding $y_ \uparrow ^{{t_1}}$, while the negative gradients of $x_ \downarrow ^{{t_1}}$ decline the corresponding $y_ \downarrow ^{{t_1}}$. Increasing the learning rate ${r_2} > {r_1}$ results in $y_ \uparrow ^{{t_1}}$ and $y_ \downarrow ^{{t_1}}$ to be pushed further out at ${t=t_1}$. When $t \ge {t_1}$, both signs of ${\raise0.7ex\hbox{${\partial l}$} \!\mathord{\left/
 {\vphantom {{\partial l} {\partial x_{ \uparrow  \downarrow }^{{t_1}{t_2}}}}}\right.\kern-\nulldelimiterspace}
\!\lower0.7ex\hbox{${\partial y_{ \uparrow  \downarrow }^{{t_1}{t_2}}}$}}$ and ${\raise0.7ex\hbox{${\partial l}$} \!\mathord{\left/
 {\vphantom {{\partial l} {\partial x_{ \downarrow  \uparrow }^{{t_1}{t_2}}}}}\right.\kern-\nulldelimiterspace}
\!\lower0.7ex\hbox{${\partial y_{ \downarrow  \uparrow }^{{t_1}{t_2}}}$}}$ are reversed but $\left| {{\raise0.7ex\hbox{${\partial {\bm{y}}}$} \!\mathord{\left/
 {\vphantom {{\partial {\bm{y}}} {\partial x_ \downarrow ^{{t_2}}}}}\right.\kern-\nulldelimiterspace}
\!\lower0.7ex\hbox{${\partial x_ \downarrow ^{{t_2}}}$}}} \right| < \left| {{\raise0.7ex\hbox{${\partial {\bm{y}}}$} \!\mathord{\left/
 {\vphantom {{\partial {\bm{y}}} {\partial x_ \uparrow ^{{t_1}}}}}\right.\kern-\nulldelimiterspace}
\!\lower0.7ex\hbox{${\partial x_ \uparrow ^{{t_1}}}$}}} \right|$ and $\left| {{\raise0.7ex\hbox{${\partial {\bm{y}}}$} \!\mathord{\left/
 {\vphantom {{\partial {\bm{y}}} {\partial x_ \uparrow ^{{t_2}}}}}\right.\kern-\nulldelimiterspace}
\!\lower0.7ex\hbox{${\partial x_ \uparrow ^{{t_2}}}$}}} \right| < \left| {{\raise0.7ex\hbox{${\partial {\bm{y}}}$} \!\mathord{\left/
 {\vphantom {{\partial {\bm{y}}} {\partial x_ \downarrow ^{{t_1}}}}}\right.\kern-\nulldelimiterspace}
\!\lower0.7ex\hbox{${\partial x_ \downarrow ^{{t_1}}}$}}} \right|$ due to Eq.(10). Therefore, it takes longer $t_2^{{r_2}} > t_2^{{r_1}}$ to restore from ${{\bm{x}}^{{t_1}}} = \left[ {x_1^{{t_1}},...,x_d^{{t_1}}} \right]$ to ${{\bm{x}}^0} = \left[ {x_1^0,...,x_d^0} \right]$. For numerical validation, we assume the simplest case where ${{\bm{x}}^0} = \left[ {0.001,0.001} \right]$, ${\raise0.7ex\hbox{${\partial l}$} \!\mathord{\left/
 {\vphantom {{\partial l} {\partial {{\bm{y}}^{{t_1}}}}}}\right.\kern-\nulldelimiterspace}
\!\lower0.7ex\hbox{${\partial {{\bm{y}}^{{t_1}}}}$}} = [{\raise0.7ex\hbox{${\partial l}$} \!\mathord{\left/
 {\vphantom {{\partial l} {\partial y_ \uparrow ^{{t_1}}}}}\right.\kern-\nulldelimiterspace}
\!\lower0.7ex\hbox{${\partial y_ \uparrow ^{{t_1}}}$}},{\raise0.7ex\hbox{${\partial l}$} \!\mathord{\left/
 {\vphantom {{\partial l} {\partial y_ \downarrow ^{{t_1}}}}}\right.\kern-\nulldelimiterspace}
\!\lower0.7ex\hbox{${\partial y_ \downarrow ^{{t_1}}}$}}] = [1, - 1]$, ${\raise0.7ex\hbox{${\partial l}$} \!\mathord{\left/
 {\vphantom {{\partial l} {\partial {{\bm{y}}^{{t_2}}}}}}\right.\kern-\nulldelimiterspace}
\!\lower0.7ex\hbox{${\partial {{\bm{y}}^{{t_2}}}}$}} = [{\raise0.7ex\hbox{${\partial l}$} \!\mathord{\left/
 {\vphantom {{\partial l} {\partial y_ \downarrow ^{{t_2}}}}}\right.\kern-\nulldelimiterspace}
\!\lower0.7ex\hbox{${\partial y_ \downarrow ^{{t_2}}}$}},{\raise0.7ex\hbox{${\partial l}$} \!\mathord{\left/
 {\vphantom {{\partial l} {\partial y_ \uparrow ^{{t_2}}}}}\right.\kern-\nulldelimiterspace}
\!\lower0.7ex\hbox{${\partial y_ \uparrow ^{{t_2}}}$}}] = [ - 1,1]$, ${t_1} = 25$ and the optimizer is SGD with momentum 0.9. Then $t_2^{{r_1}}$ is $34$ for ${r_1} = 0.001$ versus $44$ for ${r_2} = 0.01$. For a fixed training epoch $T$ such as 50 where $T = {t_1} + {t_2}$, increasing parameters learning rate will increase the impact of ${\raise0.7ex\hbox{${\partial l}$} \!\mathord{\left/
 {\vphantom {{\partial l} {\partial {A^{{t_1}}}}}}\right.\kern-\nulldelimiterspace}
\!\lower0.7ex\hbox{${\partial {A^{{t_1}}}}$}}$ thereby amplifying the updates when $t<{t_1}$ over all training phase $T$ which eventually eases the performance degradation in Exp.4. By combining the insight we got from the theoretical analysis, our observations from both warmup experiments (OS.2) and Exp.4 (OS.3), we can naturally derive the following conjecture:

\textit{\textbf{HP.2}: Well-trained weights are the non-essential condition to obtain the effective gradients of parameters in DARTS. Over-trained weights beyond a tipping point even deteriorate the performance of DARTS.}

According to HP.2, we further amplify the impact of the parameter updates in the early stage by increasing the parameter learning rate in the following experiments. To prevent from the suspicion that our hyperparameters are specially tuned, we propose a simple enough method \textbf{EX}-DARTS to \textbf{ex}change the default learning rate schemes depicted in S.1 to solve the performance degradation in DARTS-V1. This experiment can be summarized as follows:

\textit{\textbf{Exp.5}: Exchange the default learning rate schemes shown in S.1 by employing exceptional small weight learning rate (0.025$\to$0.0003) and normal parameter learning rate (0.0003$\to$0.025).}

\begin{table*}[t]
\caption{Performance comparison between EX-DARTS and other methods on NAS-BENCH-201.}
\label{table1}
\begin{center}
\begin{small}
\begin{tabular}{cccccccc}
\toprule
\multirow{2}{*}{Method} & \multirow{2}{*}{\begin{tabular}[c]{@{}c@{}}Search\\ (seconds)\end{tabular}} & \multicolumn{2}{c}{CIFAR-10} & \multicolumn{2}{c}{CIFAR-100} & \multicolumn{2}{c}{ImageNet-16-120} \\
 &  & validation & test & validation & test & validation & test \\
\midrule
RSPS \cite{li2020random} & 7587.12 & 84.16±1.69 & 87.66±1.69 & 59.00±4.60 & 58.33±4.34 & 31.56±3.28 & 31.14±3.88 \\
\textbf{EX-DARTS} & 10889.87 & 89.47±1.01 & 92.97±0.82 & 69.06±2.00 & 69.35±2.00 & 40.43±2.56 & 41.01±2.69 \\
DARTS-V1 \cite{liu2018darts} & 10889.87 & 39.77±0.00 & 54.30±0.00 & 15.03±0.00 & 15.61±0.00 & 16.43±0.00 & 16.32±0.00 \\
DARTS-V2 \cite{liu2018darts} & 29901.67 & 39.77±0.00 & 54.30±0.00 & 15.03±0.00 & 15.61±0.00 & 16.43±0.00 & 16.32±0.00 \\
GDAS \cite{dong2019searching} & 28925.91 & 90.00±0.21 & 93.51±0.13 & 71.14±0.27 & 70.61±0.26 & 41.70±1.26 & 41.84±0.90 \\
SETN \cite{dong2019one} & 31009.81 & 82.25±5.17 & 86.19±4.63 & 56.86±7.59 & 56.87±7.77 & 32.54±3.63 & 31.90±4.07 \\
ENAS \cite{cai2018efficient} & 13314.51 & 39.77±0.00 & 54.30±0.00 & 15.03±0.00 & 15.61±0.00 & 16.43±0.00 & 16.32±0.00 \\
\midrule
REA \cite{real2019regularized} & \multirow{4}{*}{N/A} & 91.19±0.31 & 93.92±0.30 & 71.81±1.12 & 71.84±0.99 & 45.15±0.89 & 45.54±1.03 \\
RS &  & 90.93±0.36 & 93.70±0.36 & 70.93±1.09 & 71.04±1.07 & 44.45±1.10 & 44.57±1.25 \\
REINFORCE &  & 91.09±0.37 & 93.85±0.37 & 71.61±1.12 & 71.71±1.09 & 45.05±1.02 & 45.24±1.18 \\
BOHB \cite{falkner2018bohb} &  & 90.82±0.53 & 93.61±0.52 & 70.74±1.29 & 70.85±1.28 & 44.26±1.36 & 44.42±1.49 \\
\midrule
ResNet & \multirow{2}{*}{N/A} & 90.83 & 93.97 & 70.42 & 70.86 & 44.53 & 43.63 \\
optimal &  & 91.61 & 94.37 & 73.49 & 73.51 & 46.77 & 47.31 \\
\bottomrule
\end{tabular}
\end{small}
\end{center}
\end{table*}

Figure~\ref{fig3}(b) illustrates the training accuracies of EX-DARTS which shows no signs of performance degradation. Table~\ref{table1} demonstrates that simply exchange the learning rate schemes is enough for DARTS-V1 to obtain the performance on par with other NAS methods under low GPU budget. NAS-BENCH-201 didn’t clearly indicate whether the same seed is used in their experiments which is particularly important for the evaluation of the gradient-based methods. In any case, our accuracies and variances come from five repeated experiments with different seeds. Comparing with GDAS, EX-DARTS has competitive performance, higher variance and one-third time overhead. Considering the effectiveness and simplicity of our method, the results of EX-DARTS in alleviating the performance degradation is impressive enough without further hyperparameter tuning especially comparing with the original DARTS-V1. Our findings from Exp.5 can be summarized as follows:

\textit{\textbf{OS.4}: Simply exchange learning rate schemes of parameters and weights can effectively solve the performance degradation and obtain competitive results on NAS-BENCH-201.}

Overfitting in machine learning is interpreted as the model learns ungeneralizable features from data. Validation set overfitting refers to that the hyperparameters tuned on the validation set cannot generalize well on the unseen data. In this case, the overfitting phenomenon is manifested as the accuracy gap between the validation and test. This phenomenon is consistent with the inherent implication of the overfitting itself under the traditional single-level one-step training paradigm, that is, we can conclude that the training suffers from overfitting when the overfitting phenomenon appears. For DARTS, if we regard the parameters as hyperparameters and the parameters training on the validation set as the hyperparameter tuning, then the performance degradation shown in the baseline in Figure~\ref{fig2}(a) perfectly matches the overfitting phenomenon of the validation set. However, DARTS is framed as a bilevel optimization and the performance evaluation is separate from the architecture search. The overfitting phenomenon can come from many non-overfitting reasons, such as search preferences \cite{pourchot2020share} or discretization discrepancy \cite{chu2020fair} which arises from the methodology of DARTS thereby does not match the actual meaning of the overfitting. We therefore conduct Exp.6 to observe whether the overfitting of the validation set can be reproduced when we delete the skip-connect from the search space.

\textit{\textbf{Exp.6}: Remove skip-connect in the search space of NAS-BENCH-201, other settings remain the same as the baseline.}

The results of Exp.6 are shown in Figure~\ref{fig3}(c) in which the performance of DARTS does not degrade on the search space without the skip-connect. This result is in line with the similar experiments conducted by \cite{chu2020fair} on DARTS search space. Our findings from Exp.6 can be summarized as follows: 

\textit{\textbf{OS.5}: Results of Exp.6 suggest that the performance degradation is correlated with some specific operations in the search space which is not as simple as a problem of the validation set overfitting.}

OS.5 also explains why the regularization of the parameters employed in Exp.2 cannot alleviate the degradation.

\subsection{Operation Selection Bias}
To learn the operation-related patterns during training of DARTS, we first introduce a concept of the magnitude of operation $o$ at epoch $t$ depicted in Eq.(11).
\begin{equation}
{m}\left(t,o \right) = \frac{{\sum\nolimits_{n = 1}^N {{{[a_o^n]}^t}} }}{N}{\rm{\ \ }}{\rm{for}}{\rm{\ }}o \in O{\rm{\ and\ }}0 < t \le T
\end{equation}
where $N$ is the number of compound edges in the cell and $O$ is the candidate operation set. ${[a_o^n]}^t$ denotes the entry within the variable distribution attached to the $o$th operation on the $n$th edge output from softmax at the epoch $t$. The operation magnitude $m(t,o)$ refers to the average strength of the operation $o$ over all $N$ compound edges in the cell during training. We plot the operation magnitude curves of Exp.5 in Figure~\ref{fig3}(d) from which we can clearly distinguish some similar operation selection patterns reoccur over the different-seed repeats. Figure~\ref{fig3}(d) shows that DARTS prefers average pool at the beginning and then it prefers 3$\times$3 convolution operation afterwards. We refer this kind of operation selection pattern as the operation selection bias or the selection preference which can be observed by keeping track of the operation magnitudes over the training epochs. By combining the results from Exp.5 and Exp.6, our findings can be summarized as follows:

\begin{figure*}[ht]
\begin{center}
\centerline{\includegraphics[width=\linewidth]{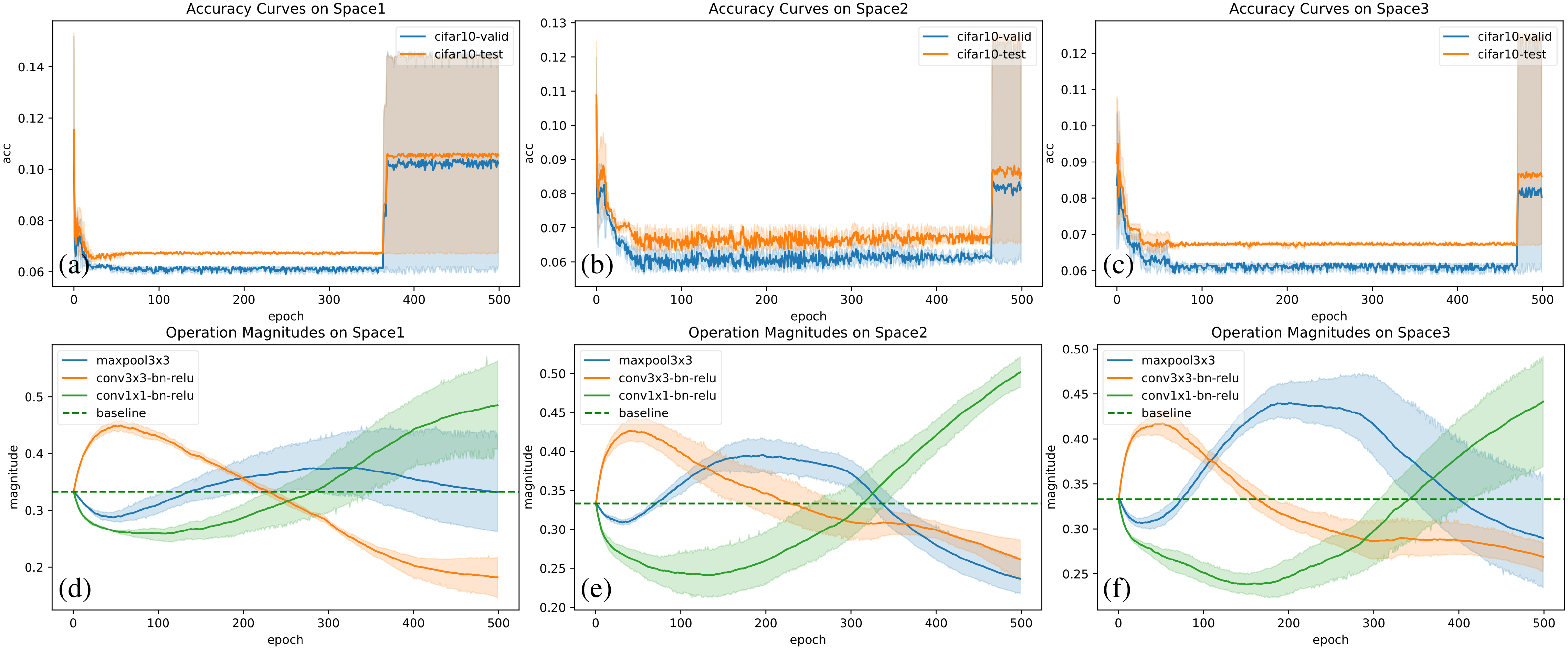}}
\caption{Accuracy and operation magnitude curves on all three search spaces in NAS-BENCH-1Shot1.}
\label{fig5}
\end{center}
\end{figure*}

\begin{figure}[ht]
\begin{center}
\centerline{\includegraphics[width=\columnwidth]{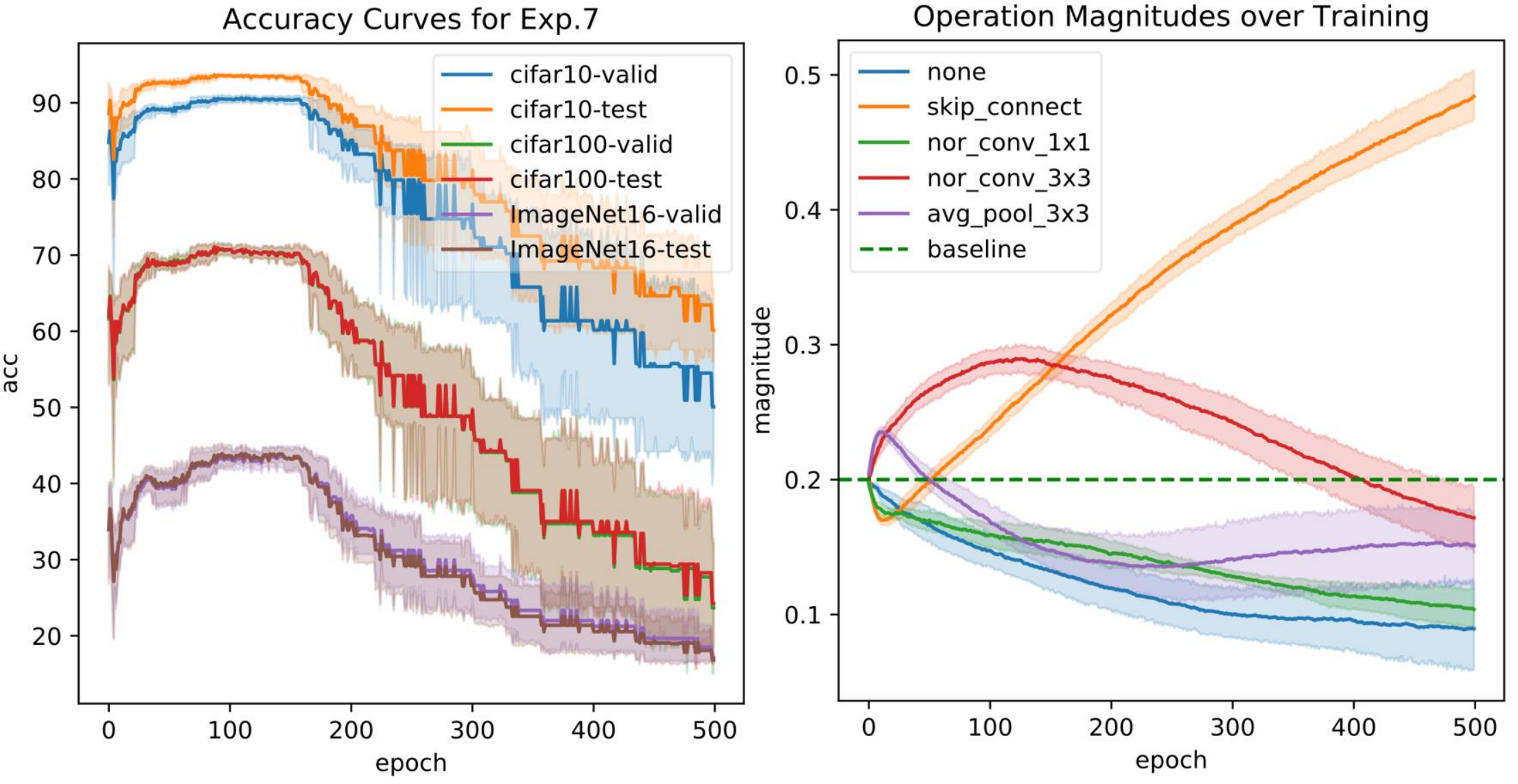}}
\caption{Accuracy and operation magnitude  curves.}
\label{fig4}
\end{center}
\end{figure}

\textit{\textbf{OS.6}: Figure~\ref{fig3}(d) demonstrates that DARTS has some operation selection biases over the bilevel optimization.}

To further validate that the OS.6 is not a coincidence but reproducible, we increase training epochs in Exp.5 and observe whether the performance degradation reoccurs. This time, we moderately reduce the parameters learning rate while keeping the exceptional small learning rate of weights to prevent the parameters from premature convergence. 

\textit{\textbf{Exp.7}: Reduce parameters learning rate moderately (0.025$\to$0.001) while keeping the weights learning rate (0.0003) and increasing the number of training epochs by 10$\times$ (50$\to$500).}

Figure~\ref{fig4} illustrates the accuracies curves (left) corresponging to the operation magnitudes during training (right) in Exp.7 from which we can find clearly that DARTS prefers average pool in the early stage of training, it prefers convolution operation in the middle stage and DARTS prefers skip-connect in the final stage which eventually leads the cell dominated by the skip-connect and results in the performance degradation. As expected, the results of Exp.7 are in line with both the selection bias in Exp.5 shown in Figure~\ref{fig3}(d) and the performance degradation in the baseline shown in Figure~\ref{fig2}(a). Our finding from Exp.7 can be summarized as:

\textit{\textbf{OS.7}: DARTS illustrates operation selection bias at different stage of training which can be reproduced across different learning rates.}

During the bilevel optimization depicted in Eq.(3), the DARTS training scheme alternately optimizes the weights and parameters with a fixed learning rate in Exp.7. One reasonable explanation of OS.7 is that the weights training affects the training of parameters thereby causing parameter optimization to prefer different operations in different training stages.  Nevertheless, OS.7 will be devalued if this selection bias is limited to a specific search space. Otherwise OS.7 is likely to be a common pattern in the bilevel optimization dynamics which has impact on all DARTS-based researches. Our speculation can be summarized as follows:

\textit{\textbf{HP.3}: OS.7 is a general pattern of the bilevel optimization in DARTS which is neither specific to the search space nor to the nasbench.}

We conduct Exp.8 to provide further empirical verification for HP.3 on three different search spaces in NAS-BENCH-1Shot1\cite{zela2019bench} which can be summarized as follows:

\textit{\textbf{Exp.8}: Keep the default hyperparameter settings and train DARTS for 500 epochs on three different search spaces in NAS-BENCH-1Shot1.}

From Figure~\ref{fig5}(d)\textasciitilde(f) we can easily distinguish some similar operation selection biases on all three search spaces. DARTS prefers 3$\times$3 convolution at first, then it gradually prefers maximum pool and finally converges to prefer 1$\times$1 convolution over the training epochs. Followings are the observation summaries from Exp.8.

\textit{\textbf{OS.8}: The operation selection bias appears on all three search spaces in NAS-BENCH-1Shot1 which is similar to OS.7 on NAS-BENCH-201;}

\textit{\textbf{OS.9}: From Figure~\ref{fig5}(a)\textasciitilde(c) we can see that DARTS also experiences mild performance degradation at the end of training on all three search spaces in NAS-BENCH-1Shot1.}

\begin{figure}[ht]
\begin{center}
\centerline{\includegraphics[width=\columnwidth]{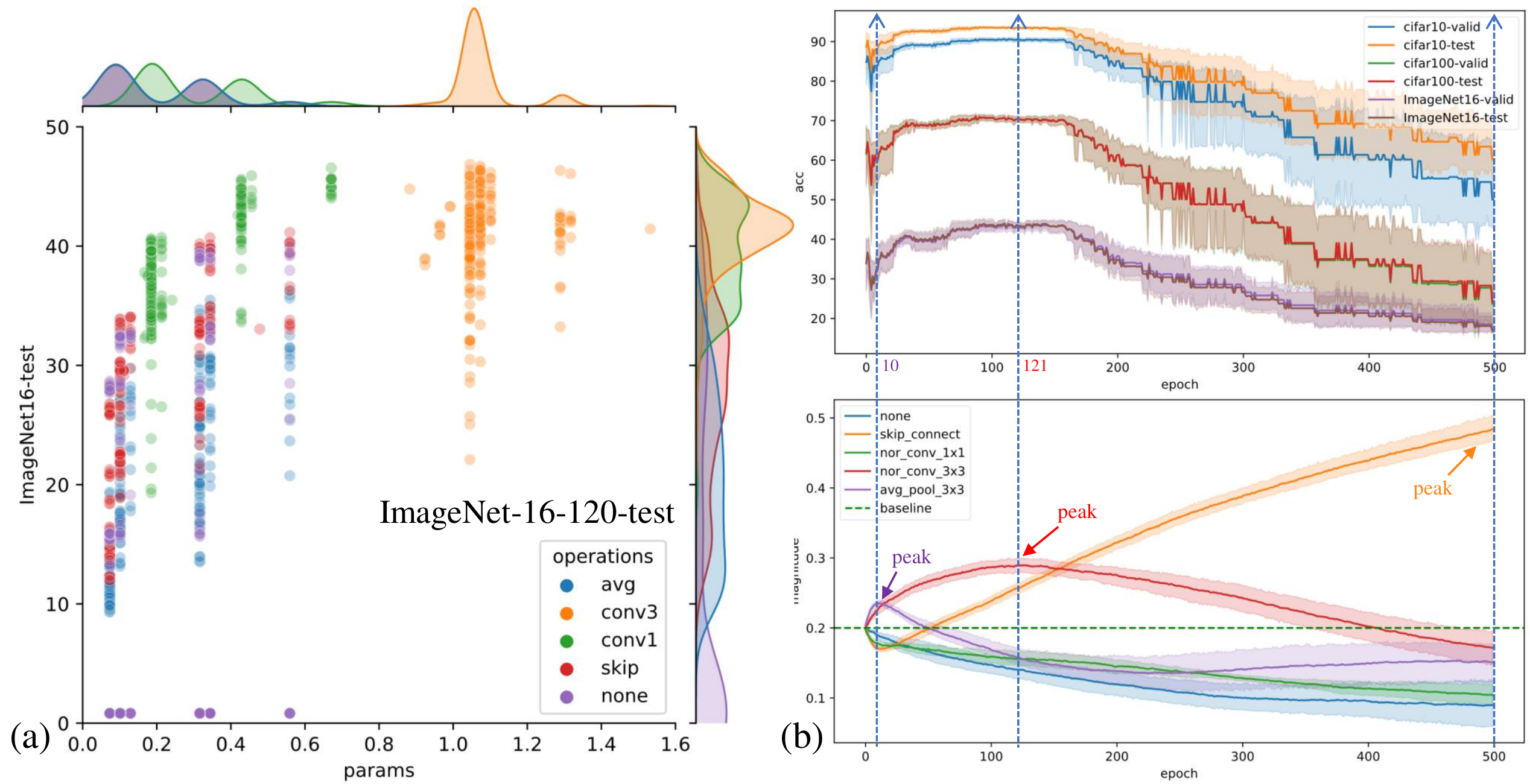}}
\caption{(a) Performance distribution on NAS-BENCH-201 when one operation occupies more than half of the candidate edges in the cell ; (b) A conceptual illustration of the selective stop on NAS-BENCH-201.}
\label{fig6}
\end{center}
\end{figure}

\subsection{Selective Stop}
We first investigate the influence of the operation selection bias on the final performance in Figure~\ref{fig6}(a). When a certain operation occupies more than half (four, five, six) of the candidate edges in the cell, we observe some performance gaps in terms of different operations in NAS-BENCH-201 which demonstrates that the operation selection bias has pronounced impact on DARTS. Figure~\ref{fig6}(a) also suggests that the performance is strongly affected by the point where DARTS eventually stop. If the training stops in the stage where the 3$\times$3 convolution is preferred, DARTS is likely to get a strong performance as shown in Exp.5. Otherwise if DARTS stops in the training stage in favor of non-parametric operations, the performance is prone to degrade like the baseline in Figure~\ref{fig2}(a). Based on this view, it naturally straightforward to propose a criterion in Eq.(12) to utilize operation magnitude to select the architecture.
\begin{equation}
\arg \mathop {\max }\limits_{A^{t}_{o}} m\left(t,o \right){\rm{\ \ for }}\ o \in O{\rm{\ and\ }}0 < t \le T
\end{equation}
where $T$ is the total epochs, Eq.(12) selects the magnitude peak of the operation $o$ as a stop criterion and takes the corresponding architecture $A^{t}_{o}$ as the final architecture. We refer the selected operation as ${o_i}$ and the operations other than ${o_i}$ as $o_{i \ne j}$ for $i,j \in [1,2,...,M]$. Eq.(12) only considers ${m({o_i})}$ which may not be robust enough in practice. We therefore formulate Eq.(13) as an alternative.
\begin{equation}
\arg \mathop {\max }\limits_{A^{t}_{o}} \sum\nolimits_{j \ne i}^M {(m(t,{o_i}){\rm{ - }}m(t,{o_j})} {\rm{)\ \ for\ }}o \in O{\rm{\ and\ }}0 < t \le T
\end{equation}
where $M$ is the number of operation candidates within each compound edge in the cell. Eq.(13) selects the peak of the sum of the residuals between the magnitudes of ${m({o_i})}$ and $m({o_{i \ne j}})$ as a stop criterion. Figure~\ref{fig6}(b) schematically depicts the usage of the stop criterion on the operation magnitude curves. Along this way, we can naturally think of a training scheme to avoid performance degradation by utilizing the operation selection bias, namely \textbf{s}elective \textbf{s}top (\textbf{SS}). The operation-magnitude-based selective stop can be summarized as Algorithm 1 and 2. Algorithm 1 uses the stop criterion $c$ for a general early stop. Algorithm 2 selects the stop points at the end of DARTS training, this way, multiple final architectures can be derived based on multiple predefined criteria $C$.

\begin{algorithm}[tb]  
   \caption{Operation-Magnitude-Based Single Point Early Stop}
   \label{alg:1}
\begin{algorithmic}
   \STATE {\bfseries Input:} Candidate operation set $O$ and stop criterion $c$
   \WHILE{not meet $c$} 
   \STATE DARTS updates (Algorithm 1 in \cite{liu2018darts})
   \STATE Calculate operation magnitude for $o \in O$
   \ENDWHILE
   \STATE Derive final architecture based on the parameters $A$
\end{algorithmic}
\end{algorithm}

\begin{algorithm}[tb]  
   \caption{Multiple Points Selective Stop}
   \label{alg:2}
\begin{algorithmic}
   \STATE {\bfseries Input:} Candidate operation set $O$, training epochs $T$, selection criteria $C$
   \WHILE{not meet $T$} 
   \STATE DARTS updates (Algorithm 1 in \cite{liu2018darts})
   \STATE Keep track of parameters and operation magnitudes for $o \in O$
   \ENDWHILE
   \FOR{$c \in C$}
   \STATE Roll back to the epoch $t$ when $c$ is met
   \STATE Derive final architecture based on the parameters in epoch $t$
   \ENDFOR
\end{algorithmic}
\end{algorithm}

\begin{table*}[t]
\caption{Evaluations on NAS-BENCH-201 on CIFAR-10.}
\label{table3}
\begin{center}
\begin{small}
\begin{tabular}{cccccccc}
\toprule
\multirow{2}{*}{Method} & \multirow{2}{*}{\begin{tabular}[c]{@{}c@{}}Search\\ (seconds)\end{tabular}} & \multicolumn{2}{c}{CIFAR-10} & \multicolumn{2}{c}{CIFAR-100} & \multicolumn{2}{c}{ImageNet-16-120} \\
 &  & validation & test & validation & test & validation & test \\
\midrule
DrNAS \cite{chen2021drnas} & 7544 & 90.15±0.10 & 93.74±0.03 & 70.82±0.27 & 71.07±0.08 & 40.76±0.05 & 41.37±0.17 \\
 &  & (91.55±0.00) & (94.36±0.00) & (73.49±0.00) & (73.51±0.00) & (46.37±0.00) & (46.34±0.00) \\
DARTS+ (sc\_2) \cite{liang2019darts+} & - & 90.68±0.64 & 93.49±0.63 & 70.36±1.46 & 70.55±1.51 & 43.85±1.50 & 43.94±1.91 \\
DARTS+ (rt\_10) \cite{liang2019darts+} & - & 58.87±21.16 & 67.76±15.47 & 32.78±20.67 & 33.21±20.43 & 21.85±7.80 & 21.33±7.43 \\
DARTS-ES \cite{arber2020understanding} & - & 84.18±14.75 & 88.23±11.40 & 62.30±16.57 & 62.55±16.52 & 38.34±8.94 & 38.32±9.41 \\
\textbf{SS-DARTS} & 7330 & \textbf{91.20±0.30} & \textbf{94.01±0.29} & \textbf{72.09±1.20} & \textbf{72.31±1.10} & \textbf{45.45±0.84} & \textbf{45.28±0.90} \\
\midrule
GAEA-bilevel \cite{li2021geometry} & 8280 & 39.77±0.00 & 54.30±0.00 & 15.03±0.00 & 15.61±0.00 & 16.43±0.00 & 16.32±0.00 \\
 &  & (80.34±6.40) & (83.08±6.15) & (53.39±8.61) & (53.92±8.77) & (28.40±6.41) & (27.11±6.48) \\
 &  & - & (91.63±2.57) & - & (68.39±4.47) & - & (41.59±4.20) \\
GAEA-bilevel (sc\_2) & - & 88.46±5.54 & 91.46±4.64 & 67.28±8.33 & 67.33±8.25 & 41.43±6.63 & 41.30±7.30 \\
\textbf{EX-GAEA-bilevel} & - & 89.94±0.27 & \textbf{93.43±0.25} & 70.56±0.42 & 70.50±0.38 & 40.13±2.28 & 40.72±2.17 \\
\textbf{SS-GAEA-bilevel} & - & \textbf{90.57±1.59} & \textbf{93.48±1.20} & \textbf{71.09±1.92} & \textbf{71.31±1.95} & \textbf{44.54±1.80} & \textbf{44.58±1.77} \\
\midrule
GAEA-ERM \cite{li2021geometry} & 14464 & 84.33±0.06 & 86.61±0.04 & 58.05±0.04 & 58.79±0.00 & 29.74±0.17 & 27.66±0.06 \\
 &  & (84.52±0.13) & (84.52±0.00) & (58.02±0.17) & (58.40±0.04) & (29.43±0.20) & (28.18±0.00) \\
 &  & - & (94.10±0.29) & - & (72.60±0.13) & - & (45.81±0.00) \\
GAEA-ERM (sc\_2) & - & 85.38±6.36 & 88.28±6.03 & 62.84±10.08 & 63.12±10.05 & 36.30±7.78 & 36.04±8.04 \\
\textbf{EX-GAEA-ERM} & - & 89.70±0.26 & \textbf{93.20±0.23} & 70.31±0.51 & 70.15±0.29 & 38.42±3.05 & 39.08±2.93 \\
\textbf{SS-GAEA-ERM} & - & \textbf{90.24±1.80} & \textbf{93.18±1.45} & \textbf{70.72±2.65} & \textbf{71.00±2.26} & \textbf{44.53±2.21} & \textbf{44.57±2.51} \\
\bottomrule
\end{tabular}
\end{small}
\end{center}
\caption{Evaluations of the weight learning rate resilience. Only the test accuracies are presented due to space limitation.}
\label{table4}
\begin{center}
\begin{small}
\begin{tabular}{cccccccccc}
\toprule
Datasets & \multicolumn{3}{c}{CIFAR-10 (test)} & \multicolumn{3}{c}{CIFAR-100 (test)} & \multicolumn{3}{c}{ImageNet-16-120 (test)} \\
learning rates & 0.003 & 0.009 & 0.015 & 0.003 & 0.009 & 0.015 & 0.003 & 0.009 & 0.015 \\
\midrule
DARTS+(sc\_2) & 93.49±0.63 & 92.89±1.16 & 91.96±3.55 & 70.56±1.51 & 69.37±2.50 & 67.54±6.11 & 43.94±1.91 & 42.39±3.70 & 41.12±5.20 \\
DARTS+(sc\_3) & 92.14±1.27 & 92.16±1.20 & 90.94±3.46 & 68.62±1.60 & 68.20±1.31 & 66.17±5.84 & 41.44±2.63 & 41.37±2.20 & 39.47±4.86 \\
DARTS+(sc\_4) & 88.82±0.95 & 86.66±8.10 & 88.76±2.95 & 62.71±2.50 & 59.57±11.3 & 62.37±5.39 & 34.62±2.95 & 32.96±9.77 & 35.62±4.48 \\
DARTS+(rt\_10) & 67.76±15.47 & 55.68±4.80 & 55.58±4.61 & 33.21±20.43 & 17.55±6.74 & 17.40±6.48 & 21.33±7.43 & 16.50±0.60 & 16.48±0.58 \\
DARTS+(rt\_2) & 93.51±0.70 & 91.04±5.07 & 86.50±8.08 & 70.81±1.84 & 67.8±7.01 & 59.95±11.91 & 43.42±2.64 & 40.20±8.32 & 34.27±8.94 \\
DARTS+(rt\_3) & 92.56±1.53 & 83.42±14.28 & 72.23±15.45 & 69.36±2.06 & 55.51±20.16 & 42.67±21.48 & 41.91±2.61 & 33.97±10.23 & 25.81±9.81 \\
DARTS+(rt\_4) & 91.78±1.80 & 74.58±16.72 & 67.73±13.80 & 67.98±2.91 & 44.04±23.72 & 33.13±18.16 & 40.25±3.59 & 27.82±12.08 & 21.53±7.17 \\
DARTS-ES & 88.23±11.40 & 79.87±13.36 & 67.15±15.03 & 62.55±16.52 & 49.69±18.75 & 32.06±19.68 & 38.32±9.40 & 29.35±8.63 & 22.10±7.70 \\
\textbf{SS-DARTS} & \textbf{94.01±0.29} & \textbf{93.77±0.40} & \textbf{93.43±0.98} & \textbf{72.31±1.20} & \textbf{71.55±1.23} & \textbf{70.92±1.81} & \textbf{45.28±0.90} & \textbf{45.10±1.31} & \textbf{44.24±2.04} \\
optimal & 94.24±0.19 & 94.03±0.32 & 93.92±0.23 & 73.02±0.83 & 72.31±1.12 & 71.61±1.04 & 46.24±0.63 & 45.59±1.21 & 45.23±1.36 \\
\bottomrule
\end{tabular}
\end{small}
\end{center}
\end{table*}

\section{Performance evaluation}
We evaluate the efficacy of our methods against multiple baselines on two search spaces: NAS-BENCH-201 and DARTS search space. We prefix \textbf{SS} and \textbf{EX} to the methods applied selective stop and exchange learning rates method respectively. All experiments are repeated five times and we report mean and standard deviation. We choose two well-developed early stop strategies DARTS+ \cite{liang2019darts+} and DARTS-ES \cite{arber2020understanding} as the baselines of the evaluation of the stop method. DARTS+ basically proposed two criteria: 1. The search procedure stops when there are two or more than two skip-connects in one normal cell (DARTS+ (sc_2)); 2. The search procedure stops when the ranking of architecture parameters for learnable operations becomes stable for 10 epochs (DARTS+ (rt_10)). We also tried to tune the hyperparameters of DARTS+ on NAS-BENCH-201 shown in Table~\ref{table4}. DARTS-ES stops training based on the dominant eigenvalues of the Hessian w.r.t the parameters. DARTS+ and SS-DARTS are evaluated in the same runs of the experiment. DARTS-ES evaluates separately while keeping all the settings unchanged except for halving the batch size due to the memory overhead of competing the Hessian. If the stop criterion is not triggered, the final architecture at the end of training (50 epochs) is used for evaluation. 

As an additional performance evaluation, we introduce two baselines GAEA \cite{li2021geometry} and DrNAS \cite{chen2021drnas} from ICLR 2021 and conduct the experiments on NAS-BENCH-201 on CIFAR-10. We notice that the experimental settings of DrNAS and GAEA are somewhat different from the standard configuration of NAS-BENCH-201 (search space, training epochs). For a fair comparison, all search methods are uniformly trained for 50 epochs on NAS-BENCH-201 standard search space (five candidate operations) which is referred to as \textbf{standard-201} and is in accordance with the code published by the original paper\footnote{https://github.com/D-X-Y/NAS-Bench-201}. Experiments of DrNAS and GAEA are based on the source codes\footnote{https://github.com/xiangning-chen/DrNAS}\footnote{https://github.com/liamcli/gaea_release} released by their authors. Only the search space and training epochs are aligned with standard-201. We provide both the performances of DrNAS and GAEA on standard-201 as well as the performances given by the original paper for clarity. The exhibit protocols of the results are as follows: 1. DrNAS: performance on standard-201 (scores come from the original paper on CIAFR-100); 2. GAEA: performance on standard-201 (scores on standard-201 except for training 25 epochs) (scores come from the original paper on the search space excluding "none"). 

For SS-DARTS, we maintain the parameter learning rate (0.001) from Exp.7 and increase the weight learning rate by 10 times (0.0003$\to$0.003) because of fewer training epochs. All the selective-stop-based (SS) methods are evaluated using algorithm 2 based on Eq.(13) after the entire 50 epochs of training. The magnitude peak of 3$\times$3 convolution is taken as the stop point except for SS-GAEA-ERM where 1$\times$1 convolution is chosen instead because of different dynamics within the single-level optimization. In the performance evaluation, the batch size of DrNAS and SS-DARTS is set to 256 to speed up the experiment, The batch size of GAEA is set to 160 due to the memory overhead.

The time cost of DARTS is always greatly affected by the environment, hyperparameter settings and implementation, which make it always difficult to normalize this part given in different papers. Therefore, we instead provide the relative time cost evaluation between SS-DARTS and the baselines in the second column in Table~\ref{table3}. By investigating the source code of the baselines, we carefully align the following aspects in the time overhead experiments: 1. Environment: Same gpu, same software version; 2. Settings: Batch size (160), init channel scale (24), training 50 epochs; 3. Implementation: Do not query the performance database. Do not evaluate supernet with the test set every epoch. Save checkpoint every epoch.

\begin{figure}[ht]
\begin{center}
\centerline{\includegraphics[width=\linewidth]{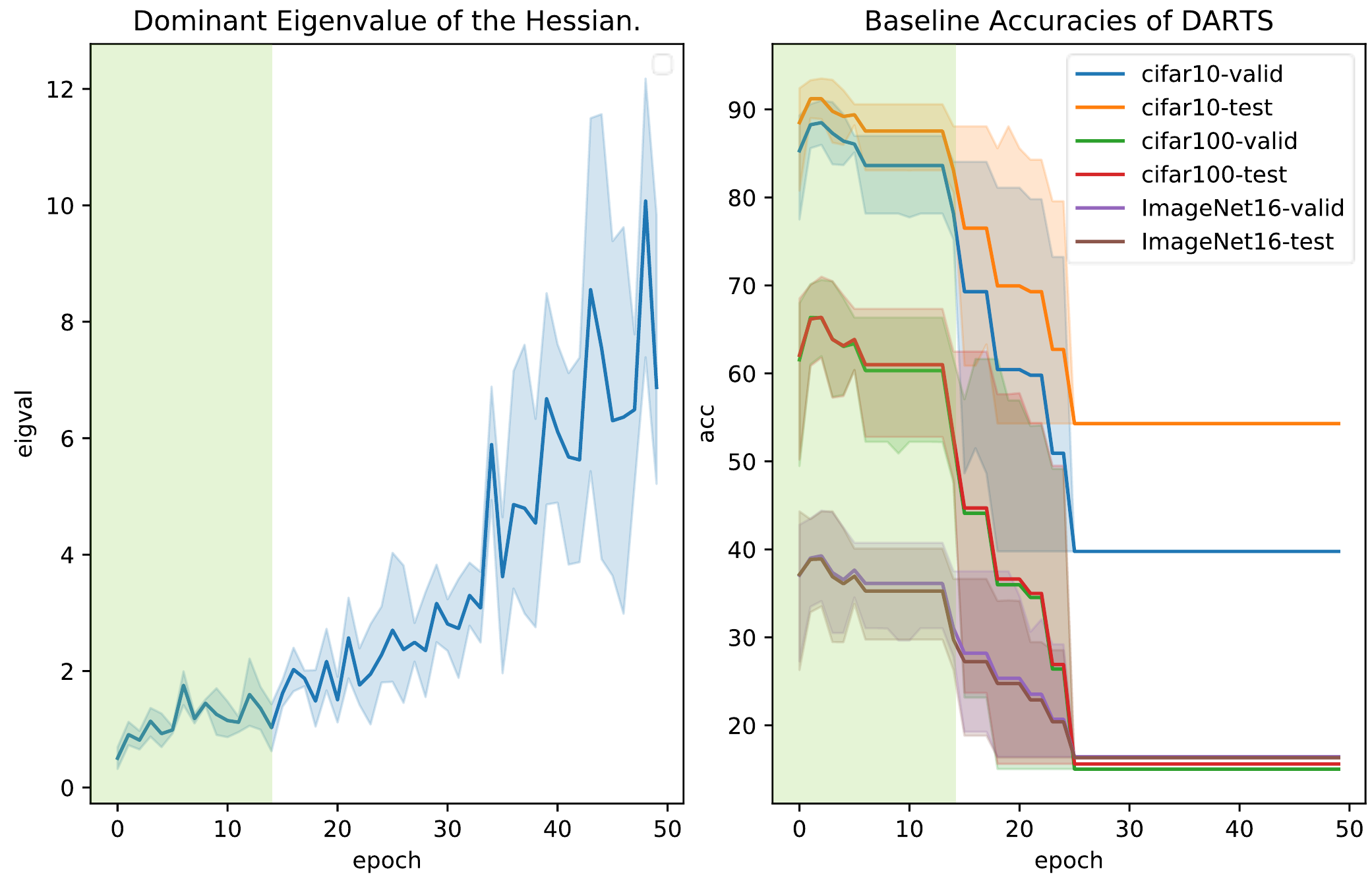}}
\caption{Searching curves corresponding to the curves of dominant eigenvalues of the Hessian w.r.t the parameters.}
\label{eigval}
\end{center}
\end{figure}

Table \ref{table3} shows all the experimental results on NAS-BENCH-201 on CIFAR-10. In the first group of Table \ref{table3}, all evaluations of the stop methods are based on DARTS-V1. We next find that GAEA (both bilevel and ERM) also suffers performance degradation on standard-201. We then employ DARTS+ (sc_2), exchange learning rates method proposed in Section 2.2 and selective stop to try to alleviate the performance degradation of GAEA. The results are presented in the second and third groups in Table \ref{table3}. In our experiments, DrNAS converges very fast and consistently end up with arch-index=1462 (six 3$\times$3 convolutions) or arch-index=138 (five 3$\times$3 convolutions and one none operation) even under different seeds. In addition, DrNas empirically prefers the operation with more parameters which can also be found in Table \ref{table5} on DARTS search space.

Stop methods are easily fragile at different weight learning rates. High weight learning rate makes it more difficult for the stop criteria to obtain an accurate stop signal. Therefore, we also compare the resilience across different weight learning rates between SS-DARTS and baselines in Table \ref{table4}. The last row of Table \ref{table4} shows the best accuracies ever obtained by DARTS which are taken as the upper bound of the performance of all stop methods. As expected shown in Table \ref{table4}, the performances of the stop methods are generally improved while decreasing the weight learning rate. DARTS-ES never shows competitive performance but incurs twice memory overhead in our experiments. The explosion of the dominant eigenvalue of the Hessian looks more like a by-product of the performance degradation rather than a nimble stop criterion which can be seen more clearly in Figure~\ref{eigval}.

\begin{table}[t]
\begin{center}
\caption{Evaluations on DARTS search space}
\label{table5}
\begin{tabular}{ccc}
\toprule
Architectures & Params (M) & Test Error (\%) \\
\midrule
DARTS-V2 \cite{liu2018darts}   &   3.3       &     2.76±0.09      \\
R-DARTS(L2) \cite{arber2020understanding} &   -      &     2.95±0.21 \\
SNAS \cite{xie2018snas} &   2.8      &     2.85±0.02 \\
PC-DARTS \cite{xu2019pc} &   3.6      &     2.57±0.07 \\
NASP \cite{yao2020efficient} &   3.3      &     2.83±0.09 \\
GAEA+PC-DARTS \cite{li2021geometry}   &   3.7       &     2.50±0.06       \\
DARTS+PT \cite{wang2020rethinking}      &  3.0         &    2.61±0.08      \\
SDARTS-RS+PT \cite{wang2020rethinking}      &  3.3         &    2.54±0.10      \\
SGAS+PT \cite{wang2020rethinking}    &    3.9       &     2.56±0.10       \\
DrNAS \cite{chen2021drnas}  &    4.0       &    2.54±0.03        \\
\midrule
SS-DARTS-SepConv5    &     3.5      &     2.54±0.06      \\
SS-DARTS-SepConv3    &     2.8      &     2.73±0.05       \\
SS-DARTS-DilConv3    &     3.0      &     2.58±0.10       \\
SS-DARTS-DilConv5    &     2.6      &     2.70±0.07      \\
\bottomrule
\end{tabular}
\end{center}
\caption{Search on CIFAR-10 and evaluate on ImageNet}
\label{table6}
\begin{center}
\begin{small}
\begin{tabular}{ccccc}
\toprule
Architectures                  & Top-1 & Top-5 & Params & GPU days \\
\midrule
DARTS-V2  \cite{liu2018darts}         & 26.7       & 8.7        & 4.7        & 1                      \\
SNAS \cite{xie2018snas}             & 27.3       & 9.2        & 4.3        & 1.5                    \\
GDAS  \cite{dong2019searching}             & 26.0       & 8.5        & 5.3        & 0.3                    \\
PC-DARTS  \cite{xu2019pc}         & 25.1       & 7.8        & 5.3        & 0.1                    \\
GAEA+PC-DARTS  \cite{li2021geometry}     & 24.3       & 7.3        & 5.6        & 0.1                    \\
\midrule
SS-DARTS-DilConv5 & 26.2       & 8.3        & 3.9        & 0.3                    \\
SS-DARTS-SepConv5  & 24.5       & 7.4        & 4.9        & 0.3 \\
\bottomrule
\end{tabular}
\end{small}
\end{center}
\end{table}

We provide the results on DARTS search space on CIFAR-10 in Table \ref{table5} as a further performance validation of the selective stop (SS). Most of our baselines \cite{chen2021drnas,li2021geometry,wang2020rethinking} from ICLR2021 are still very strong. We need to point out that DARTS does not suffer from performance degradation within 50 epochs in this search space. But Table \ref{table5} shows the versatility of our method by which we can pick out multiple architectures with different computational budget in a single run. More importantly, all the selected architectures show highly competitive performance as shown in Table \ref{table5}. Our evaluations are based on the source code from DrNAS \cite{chen2021drnas}. We keep the hyperparameter settings unchanged except for replacing the cell genotypes. Some latest research proposed the non-magnitude-based network selection method \cite{wang2020rethinking}. Their method inevitably increases the time overhead and we provide the performance comparison in Table \ref{table5}. As shown in the last row of Table \ref{table4}, the best accuracies ever obtained by DARTS are much higher than both the random search and the average performance of the search space, which suggests that the effectiveness of the magnitude may only last a short time during the training of DARTS. We leave this discussion in the future. As common practice, we also evaluate the performance on ImageNet and provide the results in Table~\ref{table6}.

\section{Conclusion}
We notice that we are not the first to study the stop criterion to alleviate the performance degradation, but we are the first to attribute the problem to a general operation selection bias and propose to obtain better performance by utilizing this bias. In our experiment, we demonstrate that even the SOTA method can still suffer performance degradation. Compared with previous stop criteria, our methods (SS and EX) have fewer hyperparameters, more robust to weight learning rate, higher accuracy and lower variance, which are validated on both DARTS and GAEA. We also show the versatility of the selective stop by which even DARTS-V1 can obtain competitive performance under low GPU budget.

\begin{acks}
The work is supported by National Natural Science Foundation of China (Grant No. 61703013 and No. 91646201) and National Key R\&D Program of China (No. 2017YFC0803300).
\end{acks}

\balance
\bibliographystyle{ACM-Reference-Format}
\bibliography{0329_final}


\begin{thebibliography}{28}

%%% the Web address.

\ifx \showCODEN    \undefined \def \showCODEN     #1{\unskip}     \fi
\ifx \showDOI      \undefined \def \showDOI       #1{#1}\fi
\ifx \showISBNx    \undefined \def \showISBNx     #1{\unskip}     \fi
\ifx \showISBNxiii \undefined \def \showISBNxiii  #1{\unskip}     \fi
\ifx \showISSN     \undefined \def \showISSN      #1{\unskip}     \fi
\ifx \showLCCN     \undefined \def \showLCCN      #1{\unskip}     \fi
\ifx \shownote     \undefined \def \shownote      #1{#1}          \fi
\ifx \showarticletitle \undefined \def \showarticletitle #1{#1}   \fi
\ifx \showURL      \undefined \def \showURL       {\relax}        \fi
% invisible to TeX
\providecommand\bibfield[2]{#2}
\providecommand\bibinfo[2]{#2}
\providecommand\natexlab[1]{#1}
\providecommand\showeprint[2][]{arXiv:#2}

\bibitem[\protect\citeauthoryear{Arber~Zela, Saikia, Marrakchi, Brox, and
  Hutter}{Arber~Zela et~al\mbox{.}}{2020}]%
        {arber2020understanding}
\bibfield{author}{\bibinfo{person}{Thomas~Elsken Arber~Zela},
  \bibinfo{person}{Tonmoy Saikia}, \bibinfo{person}{Yassine Marrakchi},
  \bibinfo{person}{Thomas Brox}, {and} \bibinfo{person}{Frank Hutter}.}
  \bibinfo{year}{2020}\natexlab{}.
\newblock \showarticletitle{Understanding and robustifying differentiable
  architecture search}. In \bibinfo{booktitle}{\emph{International Conference
  on Learning Representations}}, Vol.~\bibinfo{volume}{3}. \bibinfo{pages}{7}.
\newblock


\bibitem[\protect\citeauthoryear{Bender, Kindermans, Zoph, Vasudevan, and
  Le}{Bender et~al\mbox{.}}{2018}]%
        {bender2018understanding}
\bibfield{author}{\bibinfo{person}{Gabriel Bender}, \bibinfo{person}{Pieter-Jan
  Kindermans}, \bibinfo{person}{Barret Zoph}, \bibinfo{person}{Vijay
  Vasudevan}, {and} \bibinfo{person}{Quoc Le}.}
  \bibinfo{year}{2018}\natexlab{}.
\newblock \showarticletitle{Understanding and simplifying one-shot architecture
  search}. In \bibinfo{booktitle}{\emph{International Conference on Machine
  Learning}}. PMLR, \bibinfo{pages}{550--559}.
\newblock


\bibitem[\protect\citeauthoryear{Cai, Chen, Zhang, Yu, and Wang}{Cai
  et~al\mbox{.}}{2018}]%
        {cai2018efficient}
\bibfield{author}{\bibinfo{person}{Han Cai}, \bibinfo{person}{Tianyao Chen},
  \bibinfo{person}{Weinan Zhang}, \bibinfo{person}{Yong Yu}, {and}
  \bibinfo{person}{Jun Wang}.} \bibinfo{year}{2018}\natexlab{}.
\newblock \showarticletitle{Efficient architecture search by network
  transformation}. In \bibinfo{booktitle}{\emph{Proceedings of the AAAI
  Conference on Artificial Intelligence}}, Vol.~\bibinfo{volume}{32}.
\newblock


\bibitem[\protect\citeauthoryear{Chen, Wang, Cheng, Tang, and Hsieh}{Chen
  et~al\mbox{.}}{2021}]%
        {chen2021drnas}
\bibfield{author}{\bibinfo{person}{Xiangning Chen}, \bibinfo{person}{Ruochen
  Wang}, \bibinfo{person}{Minhao Cheng}, \bibinfo{person}{Xiaocheng Tang},
  {and} \bibinfo{person}{Cho-Jui Hsieh}.} \bibinfo{year}{2021}\natexlab{}.
\newblock \showarticletitle{DrNAS: Dirichlet Neural Architecture Search}. In
  \bibinfo{booktitle}{\emph{International Conference on Learning
  Representations}}.
\newblock


\bibitem[\protect\citeauthoryear{Chen, Xie, Wu, and Tian}{Chen
  et~al\mbox{.}}{2019}]%
        {chen2019progressive}
\bibfield{author}{\bibinfo{person}{Xin Chen}, \bibinfo{person}{Lingxi Xie},
  \bibinfo{person}{Jun Wu}, {and} \bibinfo{person}{Qi Tian}.}
  \bibinfo{year}{2019}\natexlab{}.
\newblock \showarticletitle{Progressive differentiable architecture search:
  Bridging the depth gap between search and evaluation}. In
  \bibinfo{booktitle}{\emph{Proceedings of the IEEE/CVF International
  Conference on Computer Vision}}. \bibinfo{pages}{1294--1303}.
\newblock


\bibitem[\protect\citeauthoryear{Chu, Zhou, Zhang, and Li}{Chu
  et~al\mbox{.}}{2020}]%
        {chu2020fair}
\bibfield{author}{\bibinfo{person}{Xiangxiang Chu}, \bibinfo{person}{Tianbao
  Zhou}, \bibinfo{person}{Bo Zhang}, {and} \bibinfo{person}{Jixiang Li}.}
  \bibinfo{year}{2020}\natexlab{}.
\newblock \showarticletitle{Fair darts: Eliminating unfair advantages in
  differentiable architecture search}. In \bibinfo{booktitle}{\emph{European
  Conference on Computer Vision}}. Springer, \bibinfo{pages}{465--480}.
\newblock


\bibitem[\protect\citeauthoryear{Dong and Yang}{Dong and Yang}{2019a}]%
        {dong2019bench}
\bibfield{author}{\bibinfo{person}{Xuanyi Dong} {and} \bibinfo{person}{Yi
  Yang}.} \bibinfo{year}{2019}\natexlab{a}.
\newblock \showarticletitle{NAS-Bench-201: Extending the Scope of Reproducible
  Neural Architecture Search}. In \bibinfo{booktitle}{\emph{International
  Conference on Learning Representations}}.
\newblock


\bibitem[\protect\citeauthoryear{Dong and Yang}{Dong and Yang}{2019b}]%
        {dong2019one}
\bibfield{author}{\bibinfo{person}{Xuanyi Dong} {and} \bibinfo{person}{Yi
  Yang}.} \bibinfo{year}{2019}\natexlab{b}.
\newblock \showarticletitle{One-shot neural architecture search via
  self-evaluated template network}. In \bibinfo{booktitle}{\emph{Proceedings of
  the IEEE/CVF International Conference on Computer Vision}}.
  \bibinfo{pages}{3681--3690}.
\newblock


\bibitem[\protect\citeauthoryear{Dong and Yang}{Dong and Yang}{2019c}]%
        {dong2019searching}
\bibfield{author}{\bibinfo{person}{Xuanyi Dong} {and} \bibinfo{person}{Yi
  Yang}.} \bibinfo{year}{2019}\natexlab{c}.
\newblock \showarticletitle{Searching for a robust neural architecture in four
  gpu hours}. In \bibinfo{booktitle}{\emph{Proceedings of the IEEE/CVF
  Conference on Computer Vision and Pattern Recognition}}.
  \bibinfo{pages}{1761--1770}.
\newblock


\bibitem[\protect\citeauthoryear{Falkner, Klein, and Hutter}{Falkner
  et~al\mbox{.}}{2018}]%
        {falkner2018bohb}
\bibfield{author}{\bibinfo{person}{Stefan Falkner}, \bibinfo{person}{Aaron
  Klein}, {and} \bibinfo{person}{Frank Hutter}.}
  \bibinfo{year}{2018}\natexlab{}.
\newblock \showarticletitle{BOHB: Robust and efficient hyperparameter
  optimization at scale}. In \bibinfo{booktitle}{\emph{International Conference
  on Machine Learning}}. PMLR, \bibinfo{pages}{1437--1446}.
\newblock


\bibitem[\protect\citeauthoryear{Li, Khodak, Balcan, and Talwalkar}{Li
  et~al\mbox{.}}{2021}]%
        {li2021geometry}
\bibfield{author}{\bibinfo{person}{Liam Li}, \bibinfo{person}{Mikhail Khodak},
  \bibinfo{person}{Maria-Florina Balcan}, {and} \bibinfo{person}{Ameet
  Talwalkar}.} \bibinfo{year}{2021}\natexlab{}.
\newblock \showarticletitle{Geometry-Aware Gradient Algorithms for Neural
  Architecture Search}. In \bibinfo{booktitle}{\emph{International Conference
  on Learning Representations}}.
\newblock
\urldef\tempurl%
\url{https://openreview.net/forum?id=MuSYkd1hxRP}
\showURL{%
\tempurl}


\bibitem[\protect\citeauthoryear{Li and Talwalkar}{Li and Talwalkar}{2020}]%
        {li2020random}
\bibfield{author}{\bibinfo{person}{Liam Li} {and} \bibinfo{person}{Ameet
  Talwalkar}.} \bibinfo{year}{2020}\natexlab{}.
\newblock \showarticletitle{Random search and reproducibility for neural
  architecture search}. In \bibinfo{booktitle}{\emph{Uncertainty in Artificial
  Intelligence}}. PMLR, \bibinfo{pages}{367--377}.
\newblock


\bibitem[\protect\citeauthoryear{Liang, Zhang, Sun, He, Huang, Zhuang, and
  Li}{Liang et~al\mbox{.}}{2019}]%
        {liang2019darts+}
\bibfield{author}{\bibinfo{person}{Hanwen Liang}, \bibinfo{person}{Shifeng
  Zhang}, \bibinfo{person}{Jiacheng Sun}, \bibinfo{person}{Xingqiu He},
  \bibinfo{person}{Weiran Huang}, \bibinfo{person}{Kechen Zhuang}, {and}
  \bibinfo{person}{Zhenguo Li}.} \bibinfo{year}{2019}\natexlab{}.
\newblock \showarticletitle{Darts+: Improved differentiable architecture search
  with early stopping}.
\newblock \bibinfo{journal}{\emph{arXiv preprint arXiv:1909.06035}}
  (\bibinfo{year}{2019}).
\newblock


\bibitem[\protect\citeauthoryear{Liu, Simonyan, and Yang}{Liu
  et~al\mbox{.}}{2018}]%
        {liu2018darts}
\bibfield{author}{\bibinfo{person}{Hanxiao Liu}, \bibinfo{person}{Karen
  Simonyan}, {and} \bibinfo{person}{Yiming Yang}.}
  \bibinfo{year}{2018}\natexlab{}.
\newblock \showarticletitle{DARTS: Differentiable Architecture Search}. In
  \bibinfo{booktitle}{\emph{International Conference on Learning
  Representations}}.
\newblock


\bibitem[\protect\citeauthoryear{Pourchot, Ducarouge, and Sigaud}{Pourchot
  et~al\mbox{.}}{2020}]%
        {pourchot2020share}
\bibfield{author}{\bibinfo{person}{Alo{\"\i}s Pourchot},
  \bibinfo{person}{Alexis Ducarouge}, {and} \bibinfo{person}{Olivier Sigaud}.}
  \bibinfo{year}{2020}\natexlab{}.
\newblock \showarticletitle{To share or not to share: A comprehensive appraisal
  of weight-sharing}.
\newblock \bibinfo{journal}{\emph{arXiv preprint arXiv:2002.04289}}
  (\bibinfo{year}{2020}).
\newblock


\bibitem[\protect\citeauthoryear{Real, Aggarwal, Huang, and Le}{Real
  et~al\mbox{.}}{2019}]%
        {real2019regularized}
\bibfield{author}{\bibinfo{person}{Esteban Real}, \bibinfo{person}{Alok
  Aggarwal}, \bibinfo{person}{Yanping Huang}, {and} \bibinfo{person}{Quoc~V
  Le}.} \bibinfo{year}{2019}\natexlab{}.
\newblock \showarticletitle{Regularized evolution for image classifier
  architecture search}. In \bibinfo{booktitle}{\emph{Proceedings of the aaai
  conference on artificial intelligence}}, Vol.~\bibinfo{volume}{33}.
  \bibinfo{pages}{4780--4789}.
\newblock


\bibitem[\protect\citeauthoryear{Vahdat, Mallya, Liu, and Kautz}{Vahdat
  et~al\mbox{.}}{2020}]%
        {vahdat2020unas}
\bibfield{author}{\bibinfo{person}{Arash Vahdat}, \bibinfo{person}{Arun
  Mallya}, \bibinfo{person}{Ming-Yu Liu}, {and} \bibinfo{person}{Jan Kautz}.}
  \bibinfo{year}{2020}\natexlab{}.
\newblock \showarticletitle{Unas: Differentiable architecture search meets
  reinforcement learning}. In \bibinfo{booktitle}{\emph{Proceedings of the
  IEEE/CVF Conference on Computer Vision and Pattern Recognition}}.
  \bibinfo{pages}{11266--11275}.
\newblock


\bibitem[\protect\citeauthoryear{Wang, Cheng, Chen, Tang, and Hsieh}{Wang
  et~al\mbox{.}}{2021}]%
        {wang2020rethinking}
\bibfield{author}{\bibinfo{person}{Ruochen Wang}, \bibinfo{person}{Minhao
  Cheng}, \bibinfo{person}{Xiangning Chen}, \bibinfo{person}{Xiaocheng Tang},
  {and} \bibinfo{person}{Cho-Jui Hsieh}.} \bibinfo{year}{2021}\natexlab{}.
\newblock \showarticletitle{Rethinking Architecture Selection in Differentiable
  NAS}. In \bibinfo{booktitle}{\emph{International Conference on Learning
  Representations}}.
\newblock


\bibitem[\protect\citeauthoryear{Wu, Dai, Zhang, Wang, Sun, Wu, Tian, Vajda,
  Jia, and Keutzer}{Wu et~al\mbox{.}}{2019}]%
        {wu2019fbnet}
\bibfield{author}{\bibinfo{person}{Bichen Wu}, \bibinfo{person}{Xiaoliang Dai},
  \bibinfo{person}{Peizhao Zhang}, \bibinfo{person}{Yanghan Wang},
  \bibinfo{person}{Fei Sun}, \bibinfo{person}{Yiming Wu},
  \bibinfo{person}{Yuandong Tian}, \bibinfo{person}{Peter Vajda},
  \bibinfo{person}{Yangqing Jia}, {and} \bibinfo{person}{Kurt Keutzer}.}
  \bibinfo{year}{2019}\natexlab{}.
\newblock \showarticletitle{Fbnet: Hardware-aware efficient convnet design via
  differentiable neural architecture search}. In
  \bibinfo{booktitle}{\emph{Proceedings of the IEEE/CVF Conference on Computer
  Vision and Pattern Recognition}}. \bibinfo{pages}{10734--10742}.
\newblock


\bibitem[\protect\citeauthoryear{Xie, Zheng, Liu, and Lin}{Xie
  et~al\mbox{.}}{2018}]%
        {xie2018snas}
\bibfield{author}{\bibinfo{person}{Sirui Xie}, \bibinfo{person}{Hehui Zheng},
  \bibinfo{person}{Chunxiao Liu}, {and} \bibinfo{person}{Liang Lin}.}
  \bibinfo{year}{2018}\natexlab{}.
\newblock \showarticletitle{SNAS: stochastic neural architecture search}.
\newblock \bibinfo{journal}{\emph{arXiv preprint arXiv:1812.09926}}
  (\bibinfo{year}{2018}).
\newblock


\bibitem[\protect\citeauthoryear{Xu, Xie, Zhang, Chen, Qi, Tian, and Xiong}{Xu
  et~al\mbox{.}}{2019}]%
        {xu2019pc}
\bibfield{author}{\bibinfo{person}{Yuhui Xu}, \bibinfo{person}{Lingxi Xie},
  \bibinfo{person}{Xiaopeng Zhang}, \bibinfo{person}{Xin Chen},
  \bibinfo{person}{Guo-Jun Qi}, \bibinfo{person}{Qi Tian}, {and}
  \bibinfo{person}{Hongkai Xiong}.} \bibinfo{year}{2019}\natexlab{}.
\newblock \showarticletitle{PC-DARTS: Partial Channel Connections for
  Memory-Efficient Architecture Search}. In
  \bibinfo{booktitle}{\emph{International Conference on Learning
  Representations}}.
\newblock


\bibitem[\protect\citeauthoryear{Yan, Fang, Zhang, Zheng, Zeng, Zhang, and
  Xu}{Yan et~al\mbox{.}}{2019}]%
        {yan2019hm}
\bibfield{author}{\bibinfo{person}{Shen Yan}, \bibinfo{person}{Biyi Fang},
  \bibinfo{person}{Faen Zhang}, \bibinfo{person}{Yu Zheng},
  \bibinfo{person}{Xiao Zeng}, \bibinfo{person}{Mi Zhang}, {and}
  \bibinfo{person}{Hui Xu}.} \bibinfo{year}{2019}\natexlab{}.
\newblock \showarticletitle{HM-NAS: Efficient Neural Architecture Search via
  Hierarchical Masking}. In \bibinfo{booktitle}{\emph{2019 IEEE/CVF
  International Conference on Computer Vision Workshop (ICCVW)}}. IEEE Computer
  Society, \bibinfo{pages}{1942--1950}.
\newblock


\bibitem[\protect\citeauthoryear{Yang, Esperan{\c{c}}a, and Carlucci}{Yang
  et~al\mbox{.}}{2019}]%
        {yang2019evaluation}
\bibfield{author}{\bibinfo{person}{Antoine Yang}, \bibinfo{person}{Pedro~M
  Esperan{\c{c}}a}, {and} \bibinfo{person}{Fabio~M Carlucci}.}
  \bibinfo{year}{2019}\natexlab{}.
\newblock \showarticletitle{NAS evaluation is frustratingly hard}. In
  \bibinfo{booktitle}{\emph{International Conference on Learning
  Representations}}.
\newblock


\bibitem[\protect\citeauthoryear{Yao, Xu, Tu, and Zhu}{Yao
  et~al\mbox{.}}{2020}]%
        {yao2020efficient}
\bibfield{author}{\bibinfo{person}{Quanming Yao}, \bibinfo{person}{Ju Xu},
  \bibinfo{person}{Wei-Wei Tu}, {and} \bibinfo{person}{Zhanxing Zhu}.}
  \bibinfo{year}{2020}\natexlab{}.
\newblock \showarticletitle{Efficient neural architecture search via proximal
  iterations}. In \bibinfo{booktitle}{\emph{Proceedings of the AAAI Conference
  on Artificial Intelligence}}, Vol.~\bibinfo{volume}{34}.
  \bibinfo{pages}{6664--6671}.
\newblock


\bibitem[\protect\citeauthoryear{Zela, Siems, and Hutter}{Zela
  et~al\mbox{.}}{2019}]%
        {zela2019bench}
\bibfield{author}{\bibinfo{person}{Arber Zela}, \bibinfo{person}{Julien Siems},
  {and} \bibinfo{person}{Frank Hutter}.} \bibinfo{year}{2019}\natexlab{}.
\newblock \showarticletitle{NAS-Bench-1Shot1: Benchmarking and Dissecting
  One-shot Neural Architecture Search}. In
  \bibinfo{booktitle}{\emph{International Conference on Learning
  Representations}}.
\newblock


\bibitem[\protect\citeauthoryear{Zhang, Lin, Jiang, Zhang, Wang, Xue, Zhang,
  and Yang}{Zhang et~al\mbox{.}}{2020}]%
        {zhang2020deeper}
\bibfield{author}{\bibinfo{person}{Yuge Zhang}, \bibinfo{person}{Zejun Lin},
  \bibinfo{person}{Junyang Jiang}, \bibinfo{person}{Quanlu Zhang},
  \bibinfo{person}{Yujing Wang}, \bibinfo{person}{Hui Xue},
  \bibinfo{person}{Chen Zhang}, {and} \bibinfo{person}{Yaming Yang}.}
  \bibinfo{year}{2020}\natexlab{}.
\newblock \showarticletitle{Deeper insights into weight sharing in neural
  architecture search}.
\newblock \bibinfo{journal}{\emph{arXiv preprint arXiv:2001.01431}}
  (\bibinfo{year}{2020}).
\newblock


\bibitem[\protect\citeauthoryear{Zoph and Le}{Zoph and Le}{2016}]%
        {zoph2016neural}
\bibfield{author}{\bibinfo{person}{Barret Zoph} {and} \bibinfo{person}{Quoc~V
  Le}.} \bibinfo{year}{2016}\natexlab{}.
\newblock \showarticletitle{Neural architecture search with reinforcement
  learning}.
\newblock \bibinfo{journal}{\emph{arXiv preprint arXiv:1611.01578}}
  (\bibinfo{year}{2016}).
\newblock


\bibitem[\protect\citeauthoryear{Zoph, Vasudevan, Shlens, and Le}{Zoph
  et~al\mbox{.}}{2018}]%
        {zoph2018learning}
\bibfield{author}{\bibinfo{person}{Barret Zoph}, \bibinfo{person}{Vijay
  Vasudevan}, \bibinfo{person}{Jonathon Shlens}, {and} \bibinfo{person}{Quoc~V
  Le}.} \bibinfo{year}{2018}\natexlab{}.
\newblock \showarticletitle{Learning transferable architectures for scalable
  image recognition}. In \bibinfo{booktitle}{\emph{Proceedings of the IEEE
  conference on computer vision and pattern recognition}}.
  \bibinfo{pages}{8697--8710}.
\newblock


\end{thebibliography}

\end{document}